\pdfoutput=1

\documentclass[11pt]{article}

\usepackage[]{coling}

\usepackage{times}
\usepackage{latexsym}

\usepackage[T1]{fontenc}

\usepackage[utf8]{inputenc}

\usepackage{microtype}

\usepackage{inconsolata}

\usepackage[english]{babel}
\usepackage{graphicx}
\usepackage{amsmath}
\usepackage{booktabs}
\usepackage{multirow}
\usepackage{graphicx}
\usepackage{subcaption}
\usepackage{xspace}
\usepackage[draft]{todonotes}
\usepackage{authblk}
\def\secref#1{\S\ref{sec:#1}}
\def\seclabel#1{\label{sec:#1}}

\title{How Transliterations Improve Crosslingual Alignment}

\author[1,2,*]{\bf Yihong Liu}
\author[1,2,3,*]{\bf Mingyang Wang}
\author[1,2]{\bf Amir Hossein Kargaran}
\author[1,2]{\bf Ayyoob Imani}
\author[4]{\\ \bf Orgest Xhelili}
\author[1,2]{\bf Haotian Ye}
\author[1,2]{\bf Chunlan Ma}
\author[5]{\bf François Yvon}
\author[1,2]{\bf Hinrich Sch\"utze}

\affil[1]{Center for Information and Language Processing, LMU Munich} 
\affil[2]{Munich Center for Machine Learning (MCML)} \affil[3]{Bosch Center for Artificial Intelligence}
\affil[4]{Technical University of Munich}
\affil[5]{Sorbonne Université, CNRS, ISIR, France
 \protect\\ \texttt{\{yihong, mingyang\}@cis.lmu.de}}

\begin{document}
\maketitle

\def\thefootnote{*}\footnotetext{Equal contribution.}\def\thefootnote{\arabic{footnote}}

\def\modelname{Glot500-TA\xspace}

\newcounter{notecounter}
\newcommand{\enotesoff}{\long\gdef\enote##1##2{}}
\newcommand{\enoteson}{\long\gdef\enote##1##2{{
\stepcounter{notecounter}
{\large\bf
\hspace{1cm}\arabic{notecounter} $<<<$ ##1: ##2
$>>>$\hspace{1cm}}}}}
\enoteson
\enotesoff

\begin{abstract}
    Recent studies have shown that post-aligning multilingual pretrained language models (mPLMs) using alignment objectives on both original and transliterated data can improve crosslingual alignment. This improvement further leads to better crosslingual transfer performance. However, it remains unclear how and why a better crosslingual alignment is achieved, as this technique only involves transliterations, and does not use any parallel data. This paper attempts to explicitly evaluate the crosslingual alignment and identify the key elements in transliteration-based approaches that contribute to better performance. 
    For this, we train multiple models under varying setups for two pairs of related languages: (1) \textbf{Polish} and \textbf{Ukrainian} and (2) \textbf{Hindi} and \textbf{Urdu}. To assess alignment, we define four types of similarities based on sentence representations. Our experimental results show that adding transliterations alone improves the overall similarities, even for random sentence pairs. With the help of auxiliary transliteration-based alignment objectives, especially the contrastive objective, the model learns to distinguish matched from random pairs, leading to better crosslingual alignment. However, we also show that better alignment does not always yield better downstream performance, suggesting that further research is needed to clarify the connection between alignment and performance. The code implementation is based on \url{https://github.com/cisnlp/Transliteration-PPA}.
\end{abstract}

\section{Introduction}

The training of highly multilingual language models has to cope with the diversity of scripts (e.g., more than 30 in Glot500-m \citep{imanigooghari-etal-2023-glot500}), which tends to reduce the effectiveness of crosslingual transfer \citep{dhamecha-etal-2021-role, purkayastha-etal-2023-romanization}.
A few recent studies explore the possibility of using a common script to represent all languages \citep{purkayastha-etal-2023-romanization,moosa-etal-2023-transliteration} through transliteration, a process of converting the text of a language from one script to another script \citep{wellisch1978conversion}. The intuition is that a common script can help the model to learn more knowledge through \emph{lexical overlap} since common vocabularies have been shown to contribute to better crosslinguality \citep{pires-etal-2019-multilingual,amrhein-sennrich-2020-romanization}. However, this script normalization step yields models that only support one script. This also hinders efficiency, as texts have to be transliterated into the common script before being fed to the model. In addition, \emph{transliteration ambiguity} (words with different meanings having the same transliteration) can be a potential problem for the effectiveness of crosslingual transfer \citep{liu2024transmi}.

Instead of relying solely on common-script transliterations, a recent line of work also uses transliterations as an auxiliary input to improve crosslingual alignment without expanding the vocabulary \citep{liu2024translico,xhelili-etal-2024-breaking}. These approaches combine sentences in their original script alongside their transliteration as paired inputs for sentence- or token-level alignment objectives. Surprisingly, even without parallel data, these methods show remarkable improvement in crosslingual transfer between languages with different scripts. However, the concept of crosslingual alignment is often vaguely defined in these studies. Moreover, it remains unclear why incorporating transliterations and auxiliary alignment objectives contributes to better crosslingual alignment, which relates to the similarity between translation pairs.

To this end, this work presents -- to the best of our knowledge -- the first attempt to explain why including transliterated data in training and incorporating transliteration-based alignment objectives, such as transliteration contrastive modeling (TCM) \citep{liu2024translico} and transliteration language modeling (TLM) \citep{xhelili-etal-2024-breaking}, can improve the crosslingual alignment. We first discuss definitions of crosslingual alignment \citep{roy-etal-2020-lareqa,hammerl2024understanding} and establish a clear connection between crosslingual alignment and sentence retrieval performance -- an explicit evaluation of sentence-level alignment as we show in \secref{crosslingual_alignment}. We then conduct a case study on two related language pairs using different scripts: (a) \textbf{Polish-Ukrainian} and (b) \textbf{Hindi-Urdu}, to explore how and why the transliteration-augmented approach improves crosslingual alignment. Specifically, we aim to answer the following questions:
\textbf{(1)} Does adding transliterated data alone improve crosslingual alignment?
\textbf{(2)} How do auxiliary objectives contribute to better alignment?
\textbf{(3)} How does alignment vary when the target language is in the original script, the Latin script, or when both the source and target languages are in the Latin script?
\textbf{(4)} Does better alignment always lead to better downstream zero-shot crosslingual performance?

To answer these questions, we define four types of similarities based on sentence-level representations and conduct a thorough analysis of how these similarities vary across multiple model variants throughout the pretraining stage. Our key experimental findings can be summarized as follows:

\textbf{(i)} Adding transliterations alone does not improve crosslingual alignment but does enhance all types of similarities. This occurs as the similarity between randomly paired sentences also increases. However, effective alignment requires distinguishing matched pairs from random pairs.
\textbf{(ii)}  With auxiliary transliteration-based alignment objectives, transliterations serve as an intermediary: language $L_1$ in its original script is aligned with the transliterations of $L_1$; similarly, language $L_2$ in its original script is also aligned with its transliterations; transliterations of $L_1$ are better aligned with those of $L_2$ in pretraining because of increased lexical overlap. Through this process, the alignment between $L_1$ and $L_2$ (both in their original scripts) is improved.
\textbf{(iii)} Although crosslingual alignment is generally considered crucial for enhancing zero-shot crosslingual transfer in downstream tasks, our results indicate that better alignment does not always yield better downstream performance. This finding aligns with recent findings by \citet{hua-etal-2024-mothello} and suggests that further research is needed in the community to clarify the connection between crosslingual alignment and crosslingual transfer.

\section{Related Work}

\subsection{Multilingual Language Models}

Models pretrained on a wide range of languages using self-supervised objectives, such as masked language modeling (MLM) \citep{devlin-etal-2019-bert} or causal language modeling \citep{radford2019language}, are referred to as mPLMs. 
With respect to their use of the Transformer \citep{vaswani2017attention} architecture,  these models can be categorized into encoder-only \citep{devlin-etal-2019-bert,conneau-etal-2020-unsupervised,liang-etal-2023-xlm}, encoder-decoder \citep{liu-etal-2020-multilingual-denoising,fan2021beyond,xue-etal-2021-mt5}, and decoder-only models \citep{lin-etal-2022-shot,shliazhko2022mgpt,scao2022bloom}. 
With the recent scale-up in both model and data size, decoder-only models, also known as large language models (LLMs) \citep{achiam2023gpt,touvron2023llama}, can achieve impressive performance in various generation tasks across high- and medium-resource languages \citep{zhao2024llama,ustun2024aya,zhao2024large}.
Parallel efforts have produced encoder-only models with very large language coverage, improving the situation for many low-resource or under-represented languages \citep{ogueji-etal-2021-small,alabi-etal-2022-adapting,imanigooghari-etal-2023-glot500,wang-etal-2023-nlnde,liu2023ofa}. 
These encoder-only models excel in multiple tasks in the zero-shot crosslingual transfer manner \citep{huang-etal-2019-unicoder,artetxe-schwenk-2019-massively,Hu2020xtreme,zhang-etal-2024-aadam}.

\subsection{Training with Alignment Objectives}

Most mPLMs trained without any additional crosslingual signals already show good performance across languages, possibly due to factors such as lexical overlap \citep{pires-etal-2019-multilingual}, shared position and special token embeddings \citep{dufter-schutze-2020-identifying} and even language imbalance \citep{schafer2024language}.
A recent review of factors contributing to crosslingual transfer is provided by \citet{philippy-etal-2023-towards}.
To further improve the crosslingual alignment of mPLMs, many methods additionally leverage crosslingual signals during or after pretraining. These methods can rely on bilingual dictionaries \citep{Cao2020multilingual,wu-dredze-2020-explicit,chi-etal-2021-improving,Efimov2023crosslingula}, parallel data \citep{reimers-gurevych-2020-making,pan-etal-2021-multilingual,wang-etal-2022-english}, or a combination of both \citep{wei2021universal,hu-etal-2021-explicit} to facilitate crosslingual alignment. This set of methods aims to increase the similarity between paired instances (words or sentences) and sometimes also to reduce the similarity between unrelated data, via contrastive learning objectives \citep{chopra2005learning,gao-etal-2021-simcse,chi-etal-2021-infoxlm}.
Another group of methods focuses on reformulating the training data, with the expectation of implicitly improving alignment through techniques such as artificial code-switching generation \citep{chaudhary2020dict,wang-etal-2022-expanding,reid-artetxe-2022-paradise} or using a translation language modeling objective \citep{Conneau2019xlm}.

\subsection{Transliteration in Language Modeling}

Transliteration converts the text of a language from one script to another \citep{wellisch1978conversion}. 
Since this process does not translate meanings but rather represents the original sounds as closely as possible in the target script, transliteration can be performed efficiently and accurately using a rule-based system, such as \texttt{Uroman} \citep{hermjakob-etal-2018-box}. 
Recent studies have shown that better language models can be trained using data transliterated into a common script, due to improved lexical overlap \citep{amrhein-sennrich-2020-romanization,dhamecha-etal-2021-role,muller-etal-2021-unseen,purkayastha-etal-2023-romanization,moosa-etal-2023-transliteration,ma2024exploring}.
To further break the script barrier and prevent models from supporting only one script, another line of research uses transliterations as auxiliary input to create paired data for post-pretraining with some translation-based alignment objectives \citep{liu2024translico,xhelili-etal-2024-breaking}, resulting in better crosslingual transfer performance between languages written in different scripts.
However, it remains unclear why these approaches achieve better alignment between languages written in different scripts using only transliterations as auxiliary inputs without the presence of translation data.

\section{Prelimilary: Crosslingual Alignment}\seclabel{crosslingual_alignment}

\paragraph{Definition.} \emph{Crosslingual alignment} refers to the degree of similarity among representations of similar meanings across languages, which can be further classified into \textit{weak alignment} and \textit{strong alignment} \citep{roy-etal-2020-lareqa}.\footnote{\textit{Strong alignment} requires even a greater distance of dissimilar meanings \textbf{within a language}, which is usually hard in multilingual NLP. Therefore it is not considered in this paper.} Following the definition of \textit{weak alignment}  of \citet{hammerl2024understanding}, similar meanings (across languages) should have more similar representations than dissimilar meanings. Formally, let $(u_i, v_i)$ be a pair of representations of two units (either tokens or sentences) with similar meanings in language $L_1$ and $L_2$ respectively, weak alignment is defined as follows:
\begin{equation*}
    \forall i, \text{sim}(u_i, v_i) > \max_{\forall j: j \neq i} \text{sim}(u_i, v_j)
\end{equation*}
where $\text{sim}$ is a similarity measure, such as the cosine similarity. Weak alignment requires that the representational similarity between a unit in $L_1$ and its (approximately) equivalent counterpart in $L_2$ is higher than the similarities between this unit and any other units in $L_2$. 
It is important to note that this notion of alignment emphasizes the \textbf{relative} magnitude of similarity rather than the \textbf{absolute} magnitude. The similarity between $u_i$ and $v_i$ does not have to be very large to induce compliant alignments.
Some models, though assigning similar representations to similar units, also make less related or even unrelated units similar, therefore possibly resulting in $\text{sim}(u_i, v_i) < \max_{\exists j: j \neq i} \text{sim}(u_i, v_j)$. This naturally induces suboptimal alignments, due to the failure to differentiate between similar and dissimilar meanings across languages.

\paragraph{Evaluations.} The definition of crosslingual alignment on the sentence level closely resembles the measure used in sentence retrieval tasks, where a model retrieves the most relevant or similar sentence in language $L_2$ given a query sentence in language $L_1$. Therefore, sentence-level crosslingual alignment can be \emph{directly} evaluated through sentence retrieval.
The performance of sentence retrieval can be evaluated by calculating the top-$k$ accuracy on a given parallel corpus, using sentences from one language as the queries, and retrieving their corresponding matches in the other language.\footnote{Similarly, the word alignment task (identifying corresponding words between two texts) can be used to evaluate the crosslingual alignment, particularly at the token level. However, this task often requires high-quality, golden-labeled data, which is difficult to obtain. As a result, this study focuses solely on sentence-level crosslingual alignment.}
Additionally, crosslingual alignment is believed to be able to be evaluated \emph{indirectly} by other downstream tasks that rely on zero-shot crosslingual transfer ability \citep{huang-etal-2019-unicoder,artetxe-schwenk-2019-massively}. 
That is, given an mPLM, one fine-tunes the model on the training data of a source language and then directly evaluates its performance on the test set of target languages. 
The underlying intuition is that models with strong alignment are often expected to perform well in such tasks, as representations of similar meanings should be consistent across languages.
However, we show that better crosslingual alignment does not always lead to better downstream crosslingual performance in \secref{downstream}. This observation aligns with previous findings \citep{wu-dredze-2020-explicit,gaschi-etal-2023-exploring}, which suggest that alignment and downstream performance are not always strongly correlated.

\section{Experiments \label{sec:experiments}}

\subsection{Languages}

\paragraph{Polish-Ukrainian pair.} 
Polish (pol) and Ukrainian (ukr) are Slavic languages,
belonging to the West and East Slavic branch
respectively. Polish and Ukrainian have historically
influenced each other,
contributing to shared vocabulary and linguistic features. Polish uses Latin (Latn) script while Ukrainian uses Cyrillic (Cyrl) script.

\paragraph{Hindi-Urdu pair.}
Hindi (hin) and Urdu (urd) both belong to the Indo-Aryan branch of the Indo-European family, spoken in the Indian subcontinent. They are mostly mutually intelligible languages that historically can be viewed as two standardized dialects of Hindustani, and therefore they share large common vocabularies. A major difference is that Hindi uses the Devanagari (Deva) script while Urdu uses the Arabic (Arab) script.
\\

\noindent An important difference between these two language pairs is that transliteration only changes the script of one language (ukr) for the pol-ukr pair, whereas it changes the script of both urd and hin for the hin-urd pair. In this way, our choices cover the most common cases, and therefore we assume the conclusions and insights from our experiments can be naturally extended to other language pairs.

\subsection{Training Data}

\paragraph{Original data.} We use the data from Glot500-c \citep{imanigooghari-etal-2023-glot500} for each language of interest.
For the \textbf{pol-ukr} pair, there are around 7M sentences for ukr\_Cyrl and around 19M sentences for pol\_Latn sentences.
For the \textbf{hin-urd} pair, there are around 7M sentences for hin\_Deva and 6M sentences for urd\_Arab. We concatenate all data together for each language pair and refer to the final data in their original script as \texttt{Data}$_{\text{Orig}}^{\text{pol-ukr}}$ for pol-ukr and \texttt{Data}$_{\text{Orig}}^{\text{hin-urd}}$ for hin-urd respectively.

\paragraph{Transliterated data.} We
use \texttt{Uroman} \citep{hermjakob-etal-2018-box}
to transliterate
both \texttt{Data}$_{\text{Orig}}^{\text{pol-ukr}}$
and \texttt{Data}$_{\text{Orig}}^{\text{hin-urd}}$ to the Latin script. 
We refer to the resulting Latin-script data as \texttt{Data}$_{\text{Latn}}^{\text{pol-ukr}}$ and \texttt{Data}$_{\text{Latn}}^{\text{hin-urd}}$ for pol-ukr and hin-urd pair respectively. 
It is important to note that the original and transliterated data are in one-to-one correspondence. 
This means that the $i$th line in \texttt{Data}$_{\text{Latn}}^{\text{pol-ukr}}$ is the transliteration of the $i$th line in \texttt{Data}$_{\text{Orig}}^{\text{pol-ukr}}$.

\begin{table}
\scriptsize
\centering
\setlength{\tabcolsep}{1mm}{}
\begin{tabular}{lrrrr}
\toprule
& \multicolumn{2}{c}{pol-ukr} & \multicolumn{2}{c}{hin-urd} \\
\cmidrule(lr){2-3} \cmidrule(lr){4-5}
& original & transliteration & original & transliteration \\
\midrule
\#shared token types & 2.5K & 3.9K & 2.6K & 2.3K \\
\#total token types & 21.5K & 9.6K & 24.7K & 2.4K \\
\midrule
lexical overlap & 11.6\% & 41.9\% & 10.4\% & 93.0\% \\
\bottomrule
\end{tabular}
\caption{\label{tab:lexical_overlap}
Lexical overlap between 10K randomly selected pol, ukr, hin, and urd sentences from the training data. We obtain the token types used in each language and the intersection is regarded as the shared token types. Lexical overlap is calculated as their ratio. There are many shared ones which is due to special characters and extensive code-switching. Transliterations improve lexical overlap. For hin-urd, the tokenizer only contains a small number of Latin subwords, resulting in few shared token types and total token types after transliteration.
}
\end{table}

\subsection{Training Objectives}

\paragraph{Masked Language Modeling (MLM).} This is the primary learning objective we use to train our model variants. This objective improves the general language modeling ability by masking certain tokens in the input sentences and learning to predict them. Following \citet{devlin-etal-2019-bert}, we randomly replace 15\% tokens in the input sentences with a special token: \texttt{[mask]} and use a language modeling head to reconstruct the original tokens from the final contextualized embeddings.

\paragraph{Transliteration Contrastive Modeling (TCM).} This contrastive objective is proposed by \citet{liu2024translico}. It increases the similarity between pairs of sentence-level representations composed of one sentence in its original script and the corresponding Latin transliteration. 
Following \citet{liu2024translico}, we obtain these representations by mean-pooling the output of the 8th Transformer layer and calculate the loss batch-wise: the positive samples are the paired sentences within a batch; the negative samples are any combinations of two sentences that are not paired within a batch. 

\begin{table*}[ht]
    \setlength{\belowcaptionskip}{-0.2cm}
    \scriptsize
    \centering
    \setlength{\tabcolsep}{2mm}{}
    \begin{tabular}{lrrrrrrrrrrrrrr}
        \toprule
        & \multicolumn{3}{c}{SR-B (pol $\rightarrow$ ukr)} & \multicolumn{3}{c}{SR-B (ukr $\rightarrow$ pol)} & \multicolumn{3}{c}{SR-F (pol $\rightarrow$ ukr)} & \multicolumn{3}{c}{SR-F (ukr $\rightarrow$ pol)}\\
        \cmidrule(lr){2-4} \cmidrule(lr){5-7} \cmidrule(lr){8-10} \cmidrule(lr){11-13}
        & top-1 & top-5 & top-10 & top-1 & top-5 & top-10 & top-1 & top-5 & top-10 & top-1 & top-5 & top-10\\
        \midrule
        \textbf{Model-1} & 74.7 & 88.2 & 92.2 & 74.9 & 89.1 & 92.4 & 77.3 & 91.1 & 93.5 & 78.7 & 91.4 & 94.5 \\
        \textbf{Model-2} & 70.2 & 85.3 & 88.7 & 74.7 & 90.0 & 92.7 & 74.9 & 87.8 & 91.8 & 79.7 & 91.1 & 93.7 \\
        \textbf{Model-3} & \underline{76.4} & 89.8 & \underline{92.9} & \textbf{79.8} & \textbf{92.4} & \textbf{95.1} & 75.9 & 89.9 & 94.6 & \underline{81.1} & 91.8 & 94.4 \\
        \textbf{Model-4} & 74.7 & \underline{90.0} & \underline{92.9} & 73.1 & 88.7 & 91.6 & \underline{80.3} & \underline{92.6} & \textbf{95.8} & 80.7 & \underline{92.3} & \underline{94.9} \\
        \textbf{Model-5} & \textbf{82.0} & \textbf{91.8} & \textbf{93.6} & \underline{78.2} & \underline{90.7} & \underline{93.6} & \textbf{81.6} & \textbf{92.8} & \underline{95.7} & \textbf{84.8} & \textbf{94.1} & \textbf{97.0} \\
        \midrule
        \midrule
        & \multicolumn{3}{c}{SR-B (hin $\rightarrow$ urd)} & \multicolumn{3}{c}{SR-B (urd $\rightarrow$ hin)} & \multicolumn{3}{c}{SR-F (hin $\rightarrow$ urd)} & \multicolumn{3}{c}{SR-F (urd $\rightarrow$ hin)}\\
        \cmidrule(lr){2-4} \cmidrule(lr){5-7} \cmidrule(lr){8-10} \cmidrule(lr){11-13}
        & top-1 & top-5 & top-10 & top-1 & top-5 & top-10 & top-1 & top-5 & top-10 & top-1 & top-5 & top-10\\
        \midrule
        \textbf{Model-1} & 52.7 & 71.3 & 78.2 & 44.9 & 64.0 & 74.7 & 83.5 & 94.2 & 96.1 & 81.7 & 92.5 & 95.0 \\
        \textbf{Model-2} & 50.9 & 71.8 & 79.1 & 40.0 & 59.8 & 70.4 & 84.0 & 93.4 & 95.8 & 82.7 & 93.5 & 95.5 \\
        \textbf{Model-3} & \textbf{70.2} & \textbf{82.0} & \textbf{87.6} & \textbf{77.6} & \textbf{91.1} & \textbf{93.6} & 85.2 & 94.3 & 95.7 & \underline{86.2} & \underline{95.2} & 96.2 \\
        \textbf{Model-4} & 52.9 & 72.0 & 79.3 & 42.2 & 63.1 & 72.4 & \textbf{88.2} & \textbf{95.6} & \textbf{97.1} & 85.2 & 94.6 & \underline{96.5} \\
        \textbf{Model-5} & \underline{65.1} & \underline{81.6} & \underline{86.4} & \underline{71.8} & \underline{84.4} & \underline{90.4} & \underline{88.4} & \underline{95.4} & \underline{97.0} & \textbf{86.7} & \textbf{94.7} & \textbf{96.8} \\
        \bottomrule
    \end{tabular}
    \caption{Retrieval performance. \textbf{Bold} (\underline{underlined}): best (second-best) result for each column.}
    \label{tab:case_retrieval}
\end{table*}

\paragraph{Transliteration Language Modeling (TLM).} This objective, proposed by \citet{xhelili-etal-2024-breaking}, is similar to the translation language modeling of \citet{Conneau2019xlm}, where we use transliterations, instead of translations, to build sentence pairs in the objective.
Following \citep{xhelili-etal-2024-breaking}, we concatenate a sentence and its transliteration and perform MLM on the combined text. To predict a token masked in the original sentence, the model can either attend to tokens in the original script or their transliterations, and vice versa.

\subsection{Models}

We train a SentencePiece Unigram tokenizer \citep{kudo-2018-subword,kudo-richardson-2018-sentencepiece} on \texttt{Data}$_{\text{Orig}}^{\text{pol-ukr}}$ and \texttt{Data}$_{\text{Orig}}^{\text{hin-urd}}$ for each language pair, respectively. We set the size of vocabularies to 30K for each pair. 
The tokenizers are not adapted to the transliterated data, i.e., \texttt{Data}$_{\text{Latn}}^{\text{pol-ukr}}$ and \texttt{Data}$_{\text{Latn}}^{\text{hin-urd}}$, in order to replicate the settings used by \citet{liu2024translico} and \citet{xhelili-etal-2024-breaking}, as they achieve surprisingly good performance without any tokenizer adaptation. 
As shown in Table \ref{tab:lexical_overlap}, lexical overlap increases drastically for transliterated data even without learning subwords from it. 
We then train five model variants \textbf{from scratch} for each language pair to thoroughly explore the effect of each component of the transliteration-augmented pretraining. We introduce the 5 model variants as follows (training details are reported in \secref{training_details}).\footnote{MLM is used in each model variant because it is important for language modeling. Training the models from scratch only with TCM or TLM can result in bad language modeling ability and therefore such options are not considered in our study.}

\paragraph{Model-1.} These models are trained on either \texttt{Data}$_{\text{Orig}}^{\text{pol-ukr}}$ or \texttt{Data}$_{\text{Orig}}^{\text{hin-urd}}$ only with \textbf{MLM}.

\begin{figure*}[h!]
    \setlength{\belowcaptionskip}{-0.2cm}

    \begin{subfigure}{\textwidth}
        \centering
        \includegraphics[width=\textwidth]{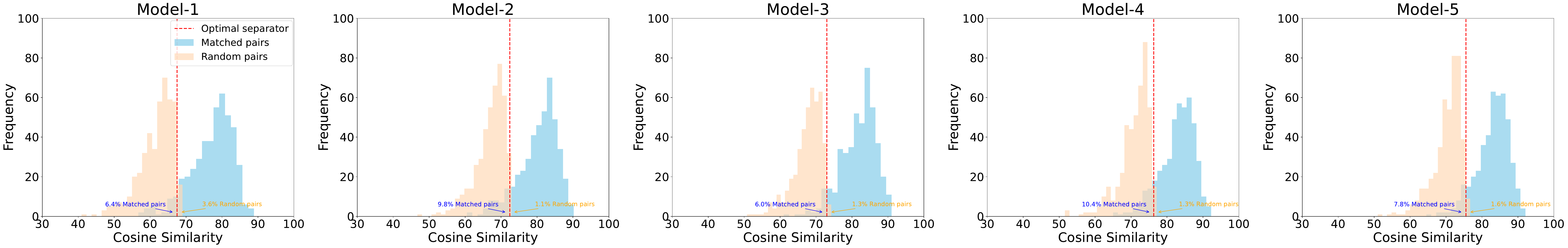}
        \caption*{Polish-Ukrainian pair: Ukrainian is $L_1$ ($s$), Polish is $L_2$ ($t$).}
    \end{subfigure}
  
  \vspace{0.5cm}

    \begin{subfigure}{\textwidth}
        \centering
        \includegraphics[width=\textwidth]{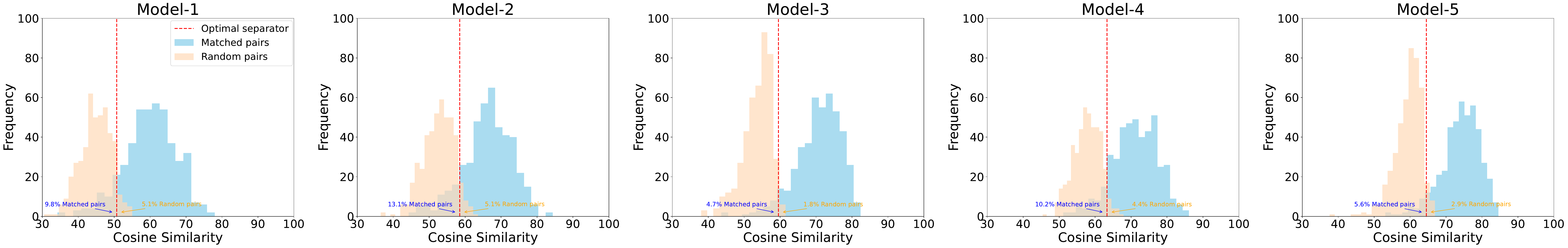}
        \caption*{(b) Hindi-Urdu pair: Urdu is $L_1$ ($s$), Hindi is $L_2$ ($t$).}
    \end{subfigure}

  \vspace{0.2cm}
  
  \caption{Histograms of similarities for matched sentence pairs and random pairs. Adding transliterated data in pretraining improves the overall similarities for both matched and random pairs (Model-2). Leveraging auxiliary objectives improves the model's ability to differentiate between matched and random sentence pairs (Model-3,-4,-5).
  }
  \label{fig:histogram_translation_sim}
\end{figure*}

\paragraph{Model-2.} These models are trained on the concatenation of the original and transliterated data only with \textbf{MLM}. For example, we concatenate \texttt{Data}$_{\text{Orig}}^{\text{pol-ukr}}$ and \texttt{Data}$_{\text{Latn}}^{\text{pol-ukr}}$ and use the resulted data as the training data for pol-ukr pair.

\paragraph{Model-3.} The training data is the same as the data used for Model-2. However, both \textbf{MLM} and \textbf{TCM} objectives are used in training. The final loss is the sum of MLM and TCM.

\paragraph{Model-4.} The training data is the same as the data used for Model-2. However, both \textbf{MLM} and \textbf{TLM} objectives are used in training. The final loss is the sum of MLM and TLM.

\paragraph{Model-5.} The training data is the same as the data used for Model-2. However, all objectives are used in training: \textbf{MLM}, \textbf{TCM}, and \textbf{TLM}. The final loss is the sum of MLM, TCM, and TLM. 

\subsection{Evaluation}\seclabel{evaluation}

\paragraph{Datasets and metric.} Since sentence retrieval directly evaluates the quality of crosslingual alignment, we focus on the sentence retrieval task as our primary evaluation. We consider two datasets: \textbf{SR-B} and \textbf{SR-F}. SR-B contains 450 parallel sentences from the Bible in each language's original script. SR-F contains 1,012 parallel sentences from Flores200 \citep{costa2022no}, also in each language's original script. We report top-1, top-5, and top-10 accuracy for each direction in each language pair.

\paragraph{Results and discussion.} Results are reported in Table~\ref{tab:case_retrieval}. We observe that Model-1 already achieves very good retrieval performance, suggesting that models can implicitly learn good crosslingual alignment even without any supervision signals, consistent with previous research findings \citep{pires-etal-2019-multilingual,dufter-schutze-2020-identifying}. 
Surprisingly, Model-2 generally performs worse than Model-1, indicating that simply adding transliterations to the training data does not improve crosslingual alignment between the two languages in their original scripts.
However, as long as any auxiliary learning objective is incorporated, retrieval performance increases. The TCM objective is particularly effective: Model-3 and Model-5 achieve the best overall retrieval performance across datasets for both pol-ukr and hin-urd pairs. 
The TLM objective is less effective compared with TCM but still helps to improve the alignment: Model-4 achieves worse performance than Model-3 but outperforms Model-1 and Model-2 in general.
Our findings can be summarized as follows: 
\textbf{(1)} vanilla MLM on related languages with different scripts can already achieve good crosslingual alignment, 
\textbf{(2)} adding transliterated data in pretraining alone has a negative impact on crosslingual alignment, and 
\textbf{(3)} alignment is improved when any auxiliary objective is included, especially TCM, which directly operates on sentence-level representations.

\begin{figure*}[h!]
    \setlength{\belowcaptionskip}{-0.1cm}

    \begin{subfigure}{\textwidth}
        \centering
        \includegraphics[width=\textwidth]{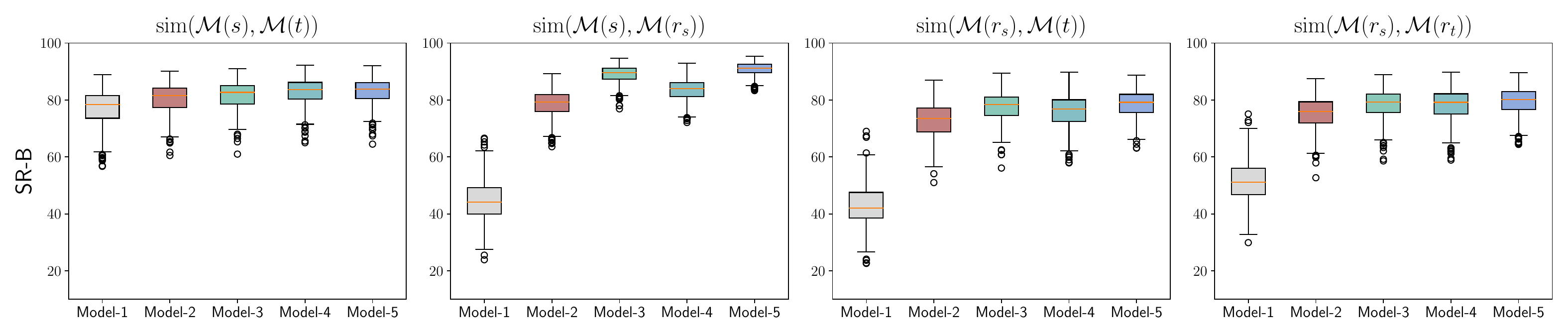}
        \caption*{(a) Polish-Ukrainian pair: Ukrainian is $L_1$ ($s$), Polish is $L_2$ ($t$).}
    \end{subfigure}
  
  \vspace{0.5cm}

    \begin{subfigure}{\textwidth}
        \centering
        \includegraphics[width=\textwidth]{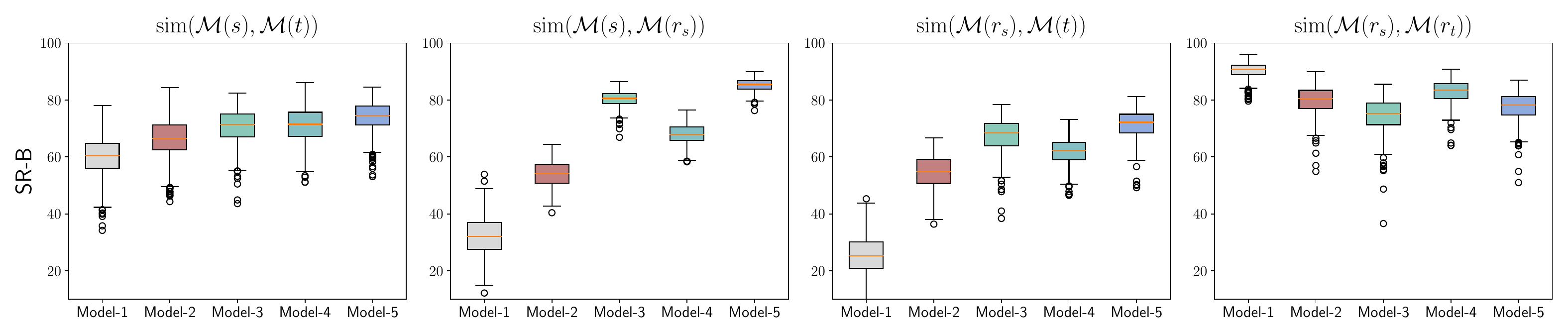}
        \caption*{(b) Hindi-Urdu pair: Urdu is $L_1$ ($s$), Hindi is $L_2$ ($t$).}
    \end{subfigure}
  
  \caption{Comparison of different types of similarities. We
  observe that the inclusion of the transliterated data not
  only improve those similarities that involve
  transliterations (i.e., $\text{sim}(\mathcal{M}(s), \mathcal{M}(r_s))$, $\text{sim}(\mathcal{M}(r_s), \mathcal{M}(t))$ and $\text{sim}(\mathcal{M}(r_s), \mathcal{M}(r_t))$), but also the similarity between the translation pairs, i.e., $\text{sim}(\mathcal{M}(s), \mathcal{M}(t))$. 
  }
  \label{fig:sim}
\end{figure*}

\section{Analysis \label{sec:analysis}}
We interpret the results of~\secref{evaluation} by
establishing a connection between crosslingual alignment and four different types of similarities
(\secref{similarity_alignment}). We also
analyze the dynamics of these similarities during the
pretraining phase (\secref{similarity_dynamics}). Finally,
we provide insights on how crosslingual alignment influences
zero-shot crosslingual transfer performance in
downstream tasks (\secref{downstream}). 
Our analysis in the following primarily focuses on
SR-B, as the impact of each component is more pronounced
(Table~\ref{tab:case_retrieval}).
See Appendix~\secref{additonal_analysis} for additional analysis on SR-F.

\subsection{Defining Similarities}\seclabel{similarity}
For a sentence $s$ written in its original script in
language $L_1$, we denote as $r_s$ its transliteration 
in Latin script, as $t$ its translation 
in language $L_2$, and as $r_t$ its transliterated translation. 
We then define the \textbf{\emph{transliteration similarity}} as $\text{sim}(\mathcal{M}(s), \mathcal{M}(r_s))$, the \textbf{\emph{translation similarity}} as $\text{sim}(\mathcal{M}(s), \mathcal{M}(t))$, the \textbf{\emph{transliteration-translation similarity}} as $\text{sim}(\mathcal{M}(r_s), \mathcal{M}(t))$, and the \textbf{\emph{transliteration-transliteration similarity}} as $\text{sim}(\mathcal{M}(r_s), \mathcal{M}(r_t))$, where $\mathcal{M}(\cdot)$ takes a text as input and encodes it as a fixed-size representation. We mean-pool the output from the 8th layer to form such fixed-size representations. For simplicity, $s$ is always ukr (resp. urd) and $t$ is always pol (resp. hin) for the pol-ukr (resp. hin-urd) pair, as both $\text{sim}(\mathcal{M}(s), \mathcal{M}(t))$ and $\text{sim}(\mathcal{M}(r_s), \mathcal{M}(r_t))$ are the same when interchanging the languages. See Appendix \secref{additonal_direction} for the other direction.

\subsection{Similarities and Alignment}\seclabel{similarity_alignment}

As discussed in \secref{crosslingual_alignment}, good crosslingual alignment does not necessarily require a model to assign high similarity to matched sentence pairs (translations). 
Instead, the model should be able to differentiate matched pairs from non-matched pairs to achieve better alignment. 
We display the similarity between matched sentence pairs and between random sentence pairs in Figure~\ref{fig:histogram_translation_sim}. We also compare the four types of similarities in each model in Figure \ref{fig:sim}.

\paragraph{Adding transliterated data alone improves similarity but not alignment.} As shown in \secref{evaluation}, simply adding transliterated data to the training data does not improve crosslingual alignment.  However, in Figure~\ref{fig:sim}, we observe that \emph{translation similarity} in Model-2 improves compared to Model-1 (from 77 to 80 in terms of the average similarity scores for the pol-ukr pair).
This suggests that the increased lexical overlap in the
added transliterated data (cf.\ Table \ref{tab:lexical_overlap}) implicitly improves overall similarities. 
Unfortunately, for this model, the similarity between random sentence pairs is also increased, as shown in Figure~\ref{fig:histogram_translation_sim}, which is detrimental to crosslingual alignments. This observation agrees with some previous studies showing that encoder-only models can mistakenly assign high cosine similarity scores to both matched and random word pairs \citep{ethayarajh-2019-contextual, zhao-etal-2021-inducing}.

\begin{figure*}
    \centering
        \setlength{\belowcaptionskip}{-0.4cm}

    \begin{subfigure}{\textwidth}
        \centering
    \includegraphics[width=0.188\textwidth]{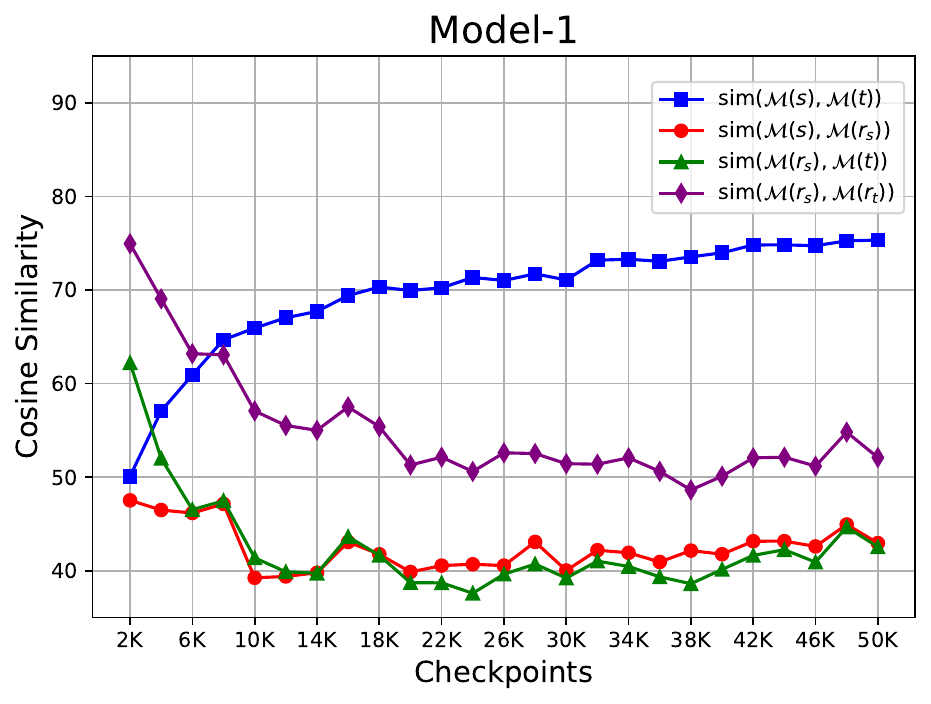}
    \includegraphics[width=0.19\textwidth]{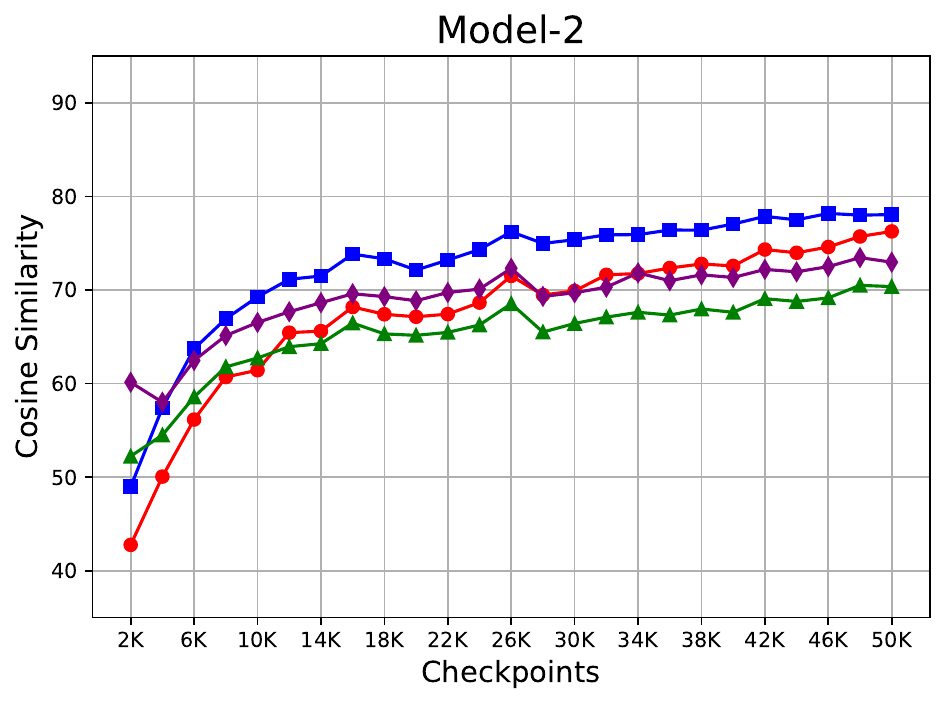}
    \includegraphics[width=0.19\textwidth]{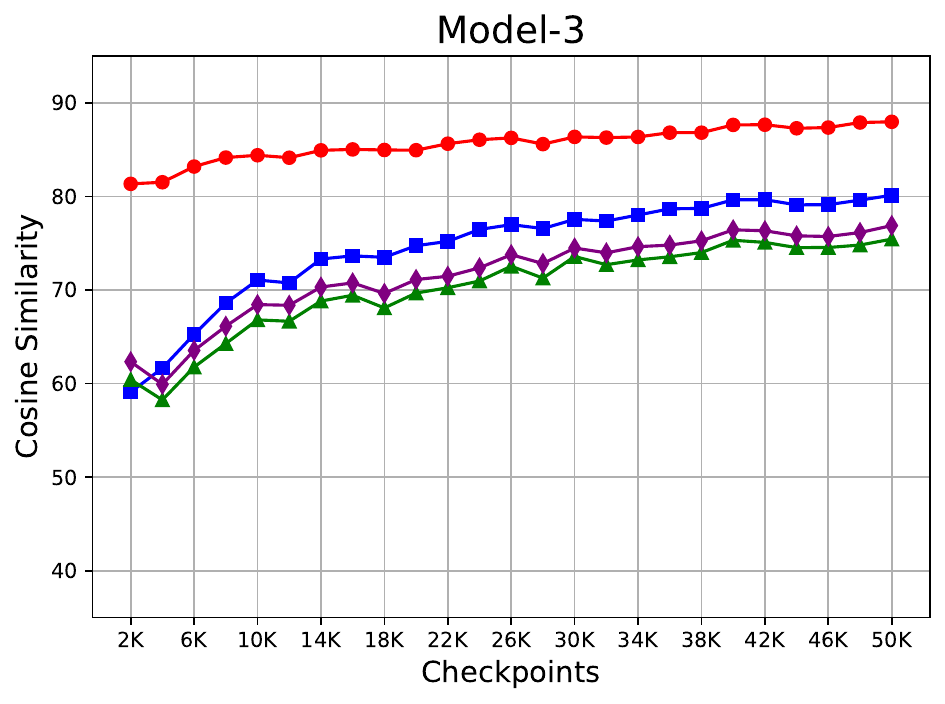}
    \includegraphics[width=0.19\textwidth]{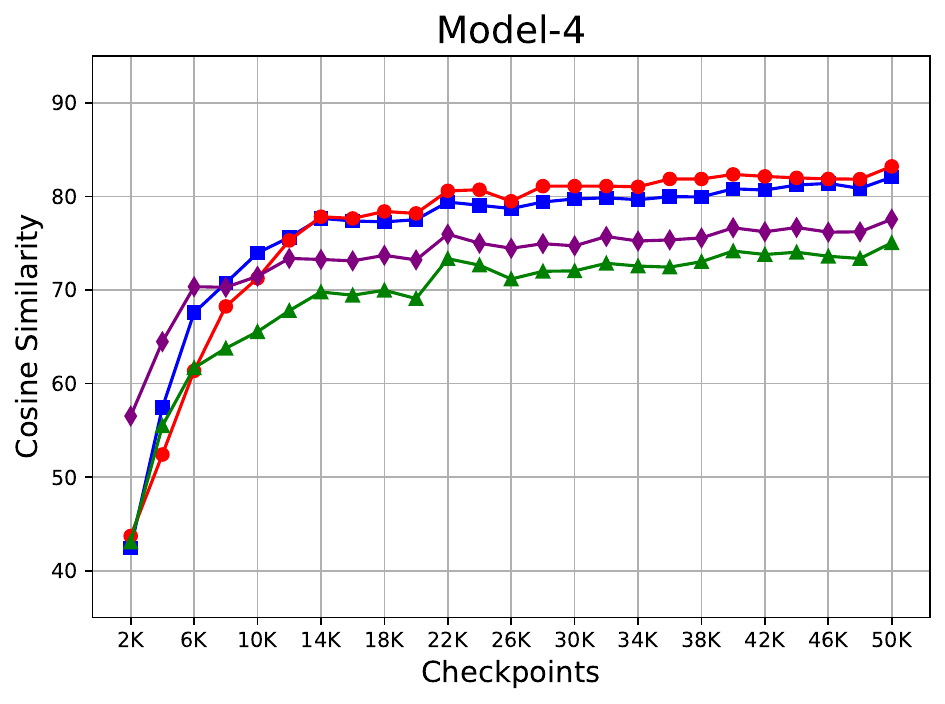}
    \includegraphics[width=0.19\textwidth]{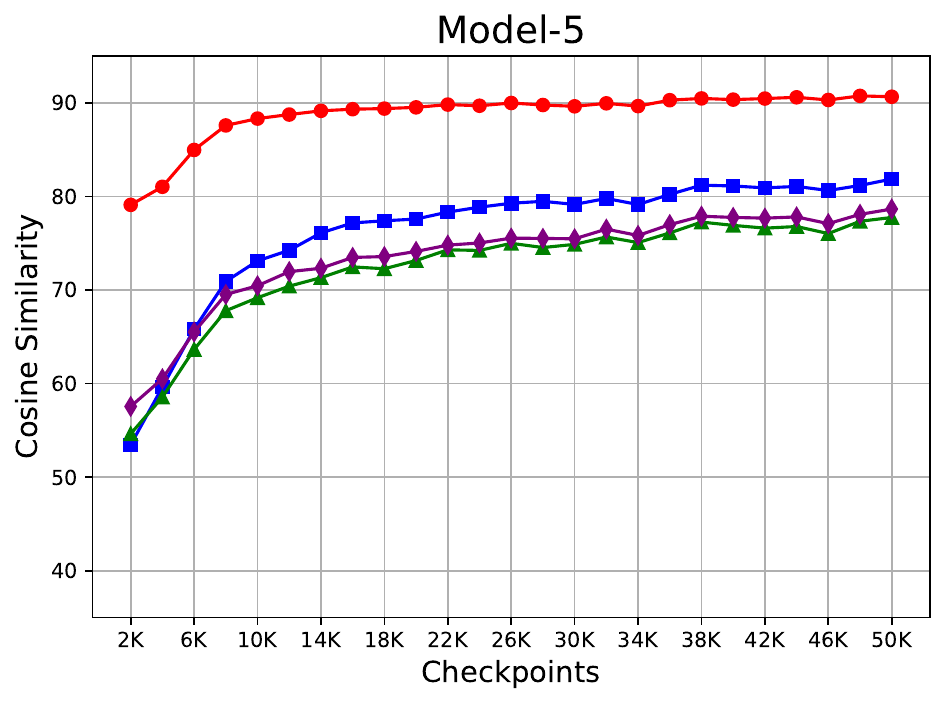}
    \caption*{(a) Polish-Ukrainian pair: Ukrainian is $L_1$ ($s$), Polish is $L_2$ ($t$).}
    \end{subfigure}
  
  \vspace{0.4cm}

    \begin{subfigure}{\textwidth}
        \centering
    \includegraphics[width=0.188\textwidth]{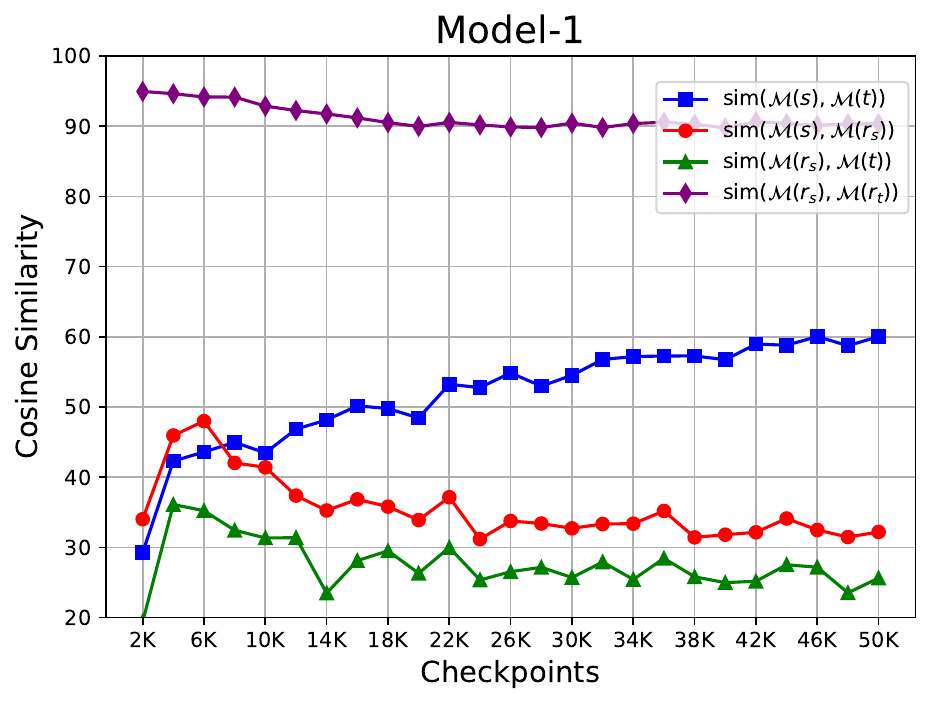}
    \includegraphics[width=0.19\textwidth]{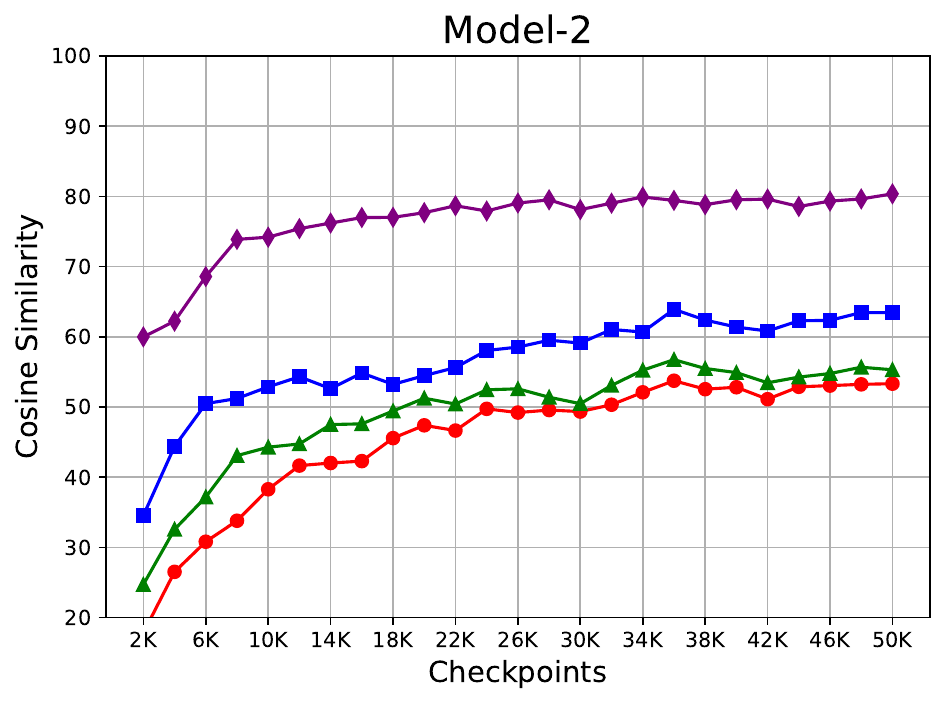}
    \includegraphics[width=0.19\textwidth]{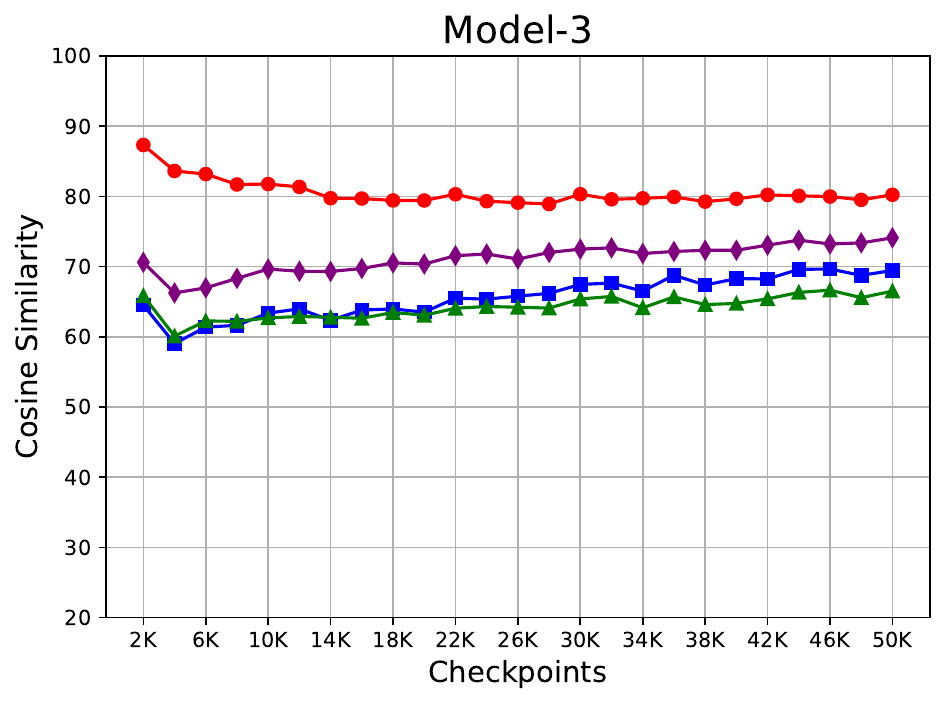}
    \includegraphics[width=0.19\textwidth]{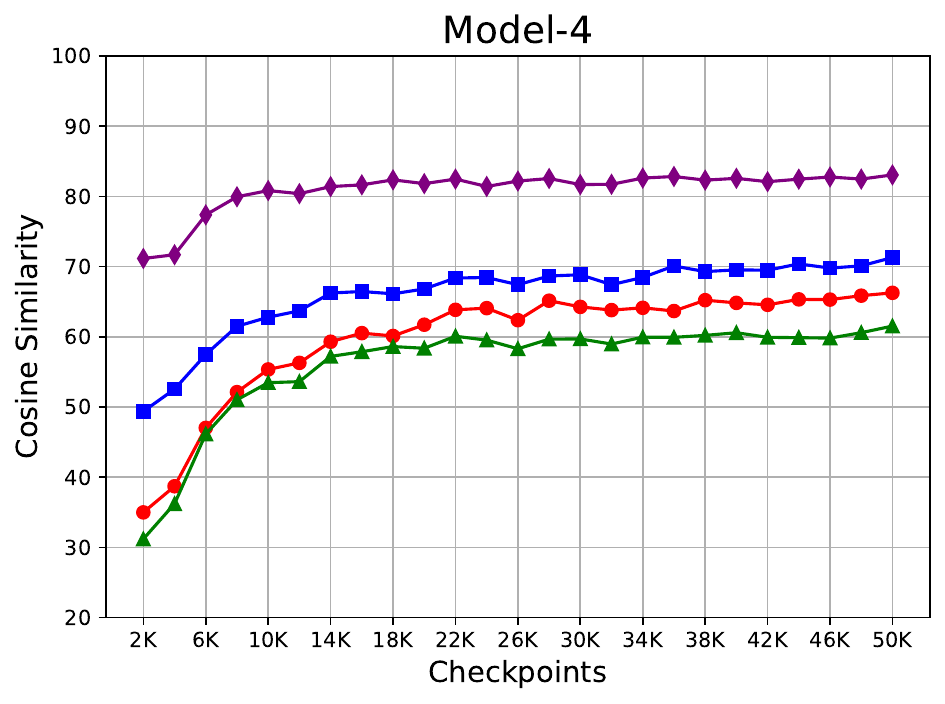}
    \includegraphics[width=0.19\textwidth]{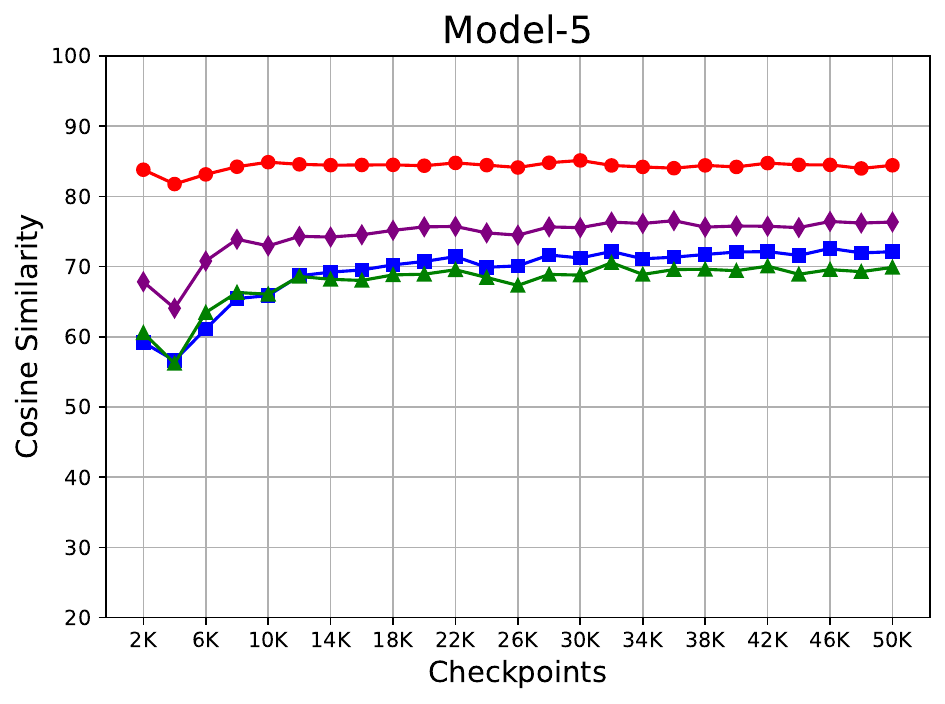}
        \caption*{(b) Hindi-Urdu pair: Urdu is $L_1$ ($s$), Hindi is $L_2$ ($t$).}
    \end{subfigure}

    \vspace{0.1cm}
    
    \caption{Dynamics of four types of similarities during training progression (from 2K to 50K checkpoints). We calculate the average of all paired sentences in SR-B for each type of similarity in each checkpoint.}
    \label{fig:progression}
\end{figure*}

\paragraph{Auxiliary learning objectives improve alignments.} Figures~\ref{fig:histogram_translation_sim} and \ref{fig:sim} show that Model-4 improves the \textit{translation similarity} -- similarity between the matched pairs, compared to Model-2, thanks to the inclusion of TLM. 
Although the similarity between random pairs also increases, the similarity gap between matched pairs and random pairs is slightly enlarged, contributing to a modest improvement in retrieval performance and crosslingual alignment. 
TCM is even more effective than TLM at improving overall similarities (see Figure~\ref{fig:sim}), while simultaneously improving the gap between matched and random pairs (see Figure~\ref{fig:histogram_translation_sim}). 
This can be attributed to the contrastive objective, which not only encourages representations of paired sentences to be similar but also teaches the model to differentiate unpaired sentences.
Consequently, we observe the best alignments in Model-3 and Model-5 for both language pairs.

\paragraph{Alignment can be improved even with a ``bad''
tokenizer.} Unlike the pol-ukr pair, where the tokenizer
already contains many Latin-script subwords due to Polish using the Latin script, the hin-urd pair does not use this
script at all.
Therefore, the tokenization results for the transliteration of Hindi or Urdu texts are ``bad'': the tokenizer often produces very long sequences composed of individual characters like ``a'' and ``b''. As a result, the overall \textit{transliteration-transliteration} similarity is very high for all model variants, as shown in Figure~\ref{fig:sim}, especially when no transliterated data is incorporated in the pretraining data (Model-1). 
However, despite such a ``bad'' tokenizer, the TCM objective significantly improves crosslingual alignment. This indicates that TCM does not necessarily rely on high-quality tokenizations of transliterated texts. In other words, its effectiveness is robust.

\subsection{Similarity Dynamics During Pretraining}\seclabel{similarity_dynamics}

To analyze the dynamics of similarities, i.e., their variation during pretraining progression, we plot all four types of similarities for each model and each language pair at every 2K steps in Figure~\ref{fig:progression}.\footnote{Each step corresponds to a single update of the parameters during pretraining. }

\paragraph{Similarities involving transliteration decrease if no transliterated data is added.} We observe high \emph{transliteration-transliteration similarity} at the early stages of the pretraining in Model-1. However, because no transliterated data is added, this similarity, along with other similarities involving transliteration, gradually drops, as shown in Figure~\ref{fig:progression}. When transliterated data is included, all transliteration-related similarities increase throughout pretraining (this effect is particularly clear when comparing Model-2 with Model-1). 
This trend can be explained by the fact that language $L_1$ in its original script and $L_1$ in the Latin script are intrinsically the same language. 
The model can quickly learn alignment between them as long as the transliterated data is included, even if explicit alignment objectives, e.g., TCM, are not used.

\paragraph{Transliterations serve as an intermediary in improving \textit{translation similarity}.} When transliterated data is included, \textit{translation similarity} increases more rapidly (as seen when comparing the similarity progression of other models with Model-1 in Figure \ref{fig:progression}). 
As all other similarities gradually decrease in Model-1, we can infer that the faster improvement in \textit{translation similarity} shown in other models is due to the improved \textit{transliteration similarity} and \textit{transliteration-transliteration similarity}, by including transliterated data.
For example, the similarity between hin\_Deva and hin\_Latn (\textit{transliteration similarity}) is improved, so is the similarity between urd\_Arab and urd\_Latn (\textit{transliteration similarity} if we refer to urd as the source language and the trend can be seen in Figure~\ref{fig:progression_sr_b_interchange}).
The improved lexical overlap in transliterated data boosts the similarity
between hin\_Latn and urd\_Latn.
The combined effect ultimately leads to further improvement in the similarity between hin\_Deva and urd\_Arab (\textit{translation-translation similarity}).
We can observe this intermediary effect is amplified when TCM is applied, as it directly optimizes the model for higher \textit{transliteration similarity} (this similarity is much higher in Model-3 and Model-5 compared to other models).

\subsection{Downstream Crosslingual Performance}\seclabel{downstream}

Although better alignment is expected to help
crosslingual transfer,
\citet{wu-dredze-2020-explicit}, \citet{gaschi-etal-2023-exploring} show
that
better token-level alignment does not always improve
performance.
We aim to further explore the connection between
the sentence-level alignment -- our focus -- and downstream
crosslingual performance. We evaluate this connection using
three datasets: SIB200 \citep{adelani-etal-2024-sib} for
text classification, WikiANN \citep{pan-etal-2017-cross} 
for named entity recognition (NER) NER, and Universal
Dependencies \citep{de-marneffe-etal-2021-universal} for
Part-of-speech tagging (POS). We report both in-language and
crosslingual transfer performance for each language pair,
with the results presented in
Table~\ref{tab:case_downstream}.

\begin{table*}[h!]
    \setlength{\belowcaptionskip}{-0.5cm}
\scriptsize
    \centering
    \setlength{\tabcolsep}{1.5mm}{}
    \begin{tabular}{lrrrr|rrrr|rrrr|}
    \toprule
            & \multicolumn{4}{c}{SIB200} & \multicolumn{4}{c}{NER} & \multicolumn{4}{c}{POS}\\
        \cmidrule(lr){2-5} \cmidrule(lr){6-9} \cmidrule(lr){10-13}
        &  \multicolumn{2}{c}{\textbf{pol}} &  \multicolumn{2}{c}{\textbf{ukr}}
        &  \multicolumn{2}{c}{\textbf{pol}} &  \multicolumn{2}{c}{\textbf{ukr}}
        &  \multicolumn{2}{c}{\textbf{pol}} &  \multicolumn{2}{c}{\textbf{ukr}}\\
        \cmidrule(lr){2-3} \cmidrule(lr){4-5}  \cmidrule(lr){6-7}  \cmidrule(lr){8-9} \cmidrule(lr){10-11} \cmidrule(lr){12-13}
        & $\rightarrow$ pol & $\rightarrow$ ukr & $\rightarrow$ ukr & $\rightarrow$ pol 
        & $\rightarrow$ pol & $\rightarrow$ ukr & $\rightarrow$ ukr & $\rightarrow$ pol 
        & $\rightarrow$ pol & $\rightarrow$ ukr & $\rightarrow$ ukr & $\rightarrow$ pol \\
        \midrule
\textbf{Model-1}	&	\textbf{80.5} & 	74.3 & 	81.0 & 	77.1 & 	85.8 & 	54.3 & 	89.8 & 	53.1 & 	98.1 & 	89.4 & 	96.8 & 	87.1 \\
\textbf{Model-2}	&	78.7 & 	74.2 & 	80.2 & 	71.8 & 	85.8 & 	\textbf{54.8} & 	90.0 & 	54.4 & 	\textbf{98.1} & 	89.7 & 	96.8 & 	86.7 \\
\textbf{Model-3}	&	75.4 & 	75.1 & 	81.0 & 	72.7 & 	86.0 & 	54.0 & 	89.7 & 	55.3 & 	98.1 & 	89.7 & 	96.9 & 	87.0 \\
\textbf{Model-4}	&	79.8 & 	76.8 & 	\textbf{83.0} & 	\textbf{78.8} & 	\textbf{86.8} & 	51.2 & 	90.1 & 	55.1 & 	98.1 & 	\textbf{90.2} & 	96.9 & 	\textbf{87.8} \\
\textbf{Model-5}	&	78.6 & 	\textbf{77.6} & 	81.7 & 	75.9 & 	86.3 & 	53.6 & 	\textbf{90.2} & 	\textbf{57.8} & 	98.1 & 	90.1 & 	\textbf{97.1} & 	87.5 \\
        \midrule
        \midrule
        &  \multicolumn{2}{c}{\textbf{hin}} &  \multicolumn{2}{c}{\textbf{urd}}
        &  \multicolumn{2}{c}{\textbf{hin}} &  \multicolumn{2}{c}{\textbf{urd}}
        &  \multicolumn{2}{c}{\textbf{hin}} &  \multicolumn{2}{c}{\textbf{urd}}\\
        \cmidrule(lr){2-3} \cmidrule(lr){4-5}  \cmidrule(lr){6-7}  \cmidrule(lr){8-9} \cmidrule(lr){10-11} \cmidrule(lr){12-13}
        & $\rightarrow$ hin & $\rightarrow$ urd & $\rightarrow$ urd & $\rightarrow$ hin 
        & $\rightarrow$ hin & $\rightarrow$ urd & $\rightarrow$ urd & $\rightarrow$ hin 
        & $\rightarrow$ hin & $\rightarrow$ urd & $\rightarrow$ urd & $\rightarrow$ hin \\
        \midrule
\textbf{Model-1}	&	\underline{82.8} & 	\underline{73.7} & 	\textbf{78.9} & 	74.2 & 	86.0 & 	31.1 & 	94.7 & 	56.5 & 	\underline{91.6} & 	83.3 & 	\textbf{92.1} & 	86.6 \\
\textbf{Model-2}	&	\textbf{83.1} & 	\textbf{75.1} & 	77.6 & 	76.3 & 	86.1 & 	36.7 & 	\underline{95.5} & 	55.0 & 	\underline{91.6} & 	\textbf{84.1} & 	\underline{92.0} & 	\underline{87.2} \\
\textbf{Model-3}	&	81.5 & 	73.1 & 	77.7 & 	\underline{76.4} & 	86.3 & 	\underline{37.0} & 	94.1 & 	58.2 & 	91.4 & 	83.4 & 	\underline{92.0} & 	87.1 \\
\textbf{Model-4}	&	82.0 & 	72.9 & 	77.1 & 	74.9 & 	\underline{86.8} & 	34.4 & 	94.9 & 	\textbf{62.0} & 	91.5 & 	\underline{83.7} & 	\underline{92.0} & 	\underline{87.2} \\
\textbf{Model-5}	&	80.4 & 	\underline{73.7} & 	\underline{78.1} & 	\textbf{79.4} & 	\textbf{87.4} & 	\textbf{40.1} & 	\textbf{95.7} & 	\underline{60.5} & 	\textbf{91.8} & 	\textbf{84.1} & 	91.9 & 	\textbf{87.4} \\
        \bottomrule
    \end{tabular}
    \caption{Downstream performance. We fine-tune each model on the training set of one language (noted with bold font), and evaluate the resulting model on both the same language (in-language evaluation, e.g., \textbf{pol} $\rightarrow$ pol) and the other language (zero-shot crosslingual transfer evaluation, e.g., \textbf{pol} $\rightarrow$ ukr). The results are averaged over three random seeds. \textbf{Bold} (\underline{underlined}): best (second-best) result for each column.}
    \label{tab:case_downstream}
\end{table*}

\paragraph{Auxiliary objectives can be detrimental for in-language evaluation but beneficial for transfer.} We observe a decrease in performance when training and evaluating on the same language. 
For instance, for SIB200, when training on Hindi
and evaluating on Hindi, Model-3, 4 and 5 are
worse than Model-1.  Similar trends are observed for Polish
and Urdu. Conversely, when training on one language and
evaluating on another, there is often a performance
improvement.  This suggests that auxiliary objectives may
negatively impact the quality of the representations within
a specific language, resulting in worse in-language
performance. However, better alignment enhances the
similarity of representations for similar sentences across
languages, which can be beneficial for zero-shot
crosslingual transfer.

\paragraph{Better crosslingual alignment does not always
improve transfer.} Although Model-3, 4, and 5 demonstrate
better alignment compared to Model-1 (cf.\ Table \ref{tab:case_retrieval}), the crosslingual
transfer performance does not substantially improve
-- especially when considering the magnitude of
alignment improvement seen in
Table~\ref{tab:case_retrieval}. This is particularly clear
in sequential tasks like NER and POS, where all models
achieve comparable performance, regardless of whether the
evaluation is in-language or crosslingual. Even for SIB200,
the improvement in Model-3, 4 and 5 is inconsistent: there
is much better transfer performance for the directions
$\text{pol} \rightarrow \text{ukr}$ and
$\text{urd}
 \rightarrow \text{hin}$
but slightly worse performance for the directions $\text{ukr} \rightarrow \text{pol}$ and $\text{hin} \rightarrow \text{urd}$. Therefore, our results suggest that better sentence-level crosslingual alignments do not consistently lead to improved crosslingual transfer, especially for sequential tasks such as NER and POS. 
We conjecture that the lack of explicit \emph{token-level alignment objectives} with word-level aligned data in our models might explain why we do not see improvements in these tasks, similar to the findings from \citet{chaudhary2020dict} and \citet{xhelili-etal-2024-breaking}.

\section{Conclusion}\seclabel{conclusion}
Our work presents the first in-depth study exploring why and how transliterations contribute to better crosslingual alignment.
We show that
adding transliterated data can improve crosslingual
alignment as transliteration acts as an
intermediary between pairs of mutual translations.
This effect is particularly pronounced when
auxiliary alignment objectives are applied, allowing models
to better distinguish matched pairs from random pairs,
thereby improving the overall alignment.
However, our empirical results also show that improved
alignment does not consistently produce better downstream performance, suggesting more research is needed to better understand the relationship between crosslingual alignment and crosslingual transfer.


\section{Future Work}
We see possible future work to overcome the limitations mentioned in the Limitations Section. A possible direction to expand this work is to explore more language pairs, or even involve more than two related languages in the training to investigate the effect of transliteration in a highly multilingual context. 
For further assessing token-level alignment, one possible way is to use word-level aligned data, which is unfortunately not much in the community. As an alternative, one can use (round-trip) word alignments, which have been shown to be very hard to find \citep{imanigooghari-etal-2023-glot500}.

\section*{Limitations}\seclabel{limitations}
This work presents the first attempt to explain why the transliteration-augmented methods can improve crosslingual alignment, which usually requires parallel data in the training or fine-tuning. One possible limitation is the number of language pairs we consider: we only use two language pairs, each of which contains two related languages that use different scripts.
Another possible limitation is that we only focus on the sentence-level crosslingual alignment in this paper and do not discuss the token-level crosslingual alignment.

\section*{Acknowledgements}

This research was supported by DFG (grant SCHU 2246/14-1)
and The European Research Council (NonSequeToR, grant \#740516).

\bibliography{custom}

\appendix

\section{Training Details}\seclabel{training_details}
To pretrain different model variants for each language pair, we use the AdamW optimizer \citep{ba2015adam,loshchilov2018decoupled} with $(\beta_1, \beta_2) = (0.9, 0.999)$ and $\epsilon = \text{1e-6}$. 
The initial learning rate is set to 5e-5. The effective batch size is 1,024 in each training step, with gradient accumulation set to 16 and 8 training instances (each instance contains a pair of sentences, see paragraph below for explanation) are used for each of the 8 NVIDIA RTX 2080Ti GPUs ($8 \times 8 \times 16 = 1,024$).
We use FP16 training with mixed precision \citep{paulius2018mixed}. We store checkpoints every 2K steps and apply early stopping based on the best average performance on SR-B retrieval task. The pretraining takes around 2 days for each model.

Except for Model-1, all other models double the training data due to the inclusion of transliterated data. This can result in a different number of parameter updates in an epoch between Model-1 and other models (the hyperparameters used by the AdamW optimizer will be different for each step), adding confounding variables to our analysis. To solve this problem, every instance in each batch is a pair of sentences in the pretraining. For Model-1, two identical sentences (in the original script) are used to form a pair, whereas a sentence and its transliteration are used to form a pair in other models. This setup ensures the total training steps in an epoch are the same for all models.

\section{Additional Analysis on SR-F}\seclabel{additonal_analysis}

\begin{figure*}
    \setlength{\belowcaptionskip}{-0.5cm}

    \begin{subfigure}{\textwidth}
        \centering
        \includegraphics[width=\textwidth]{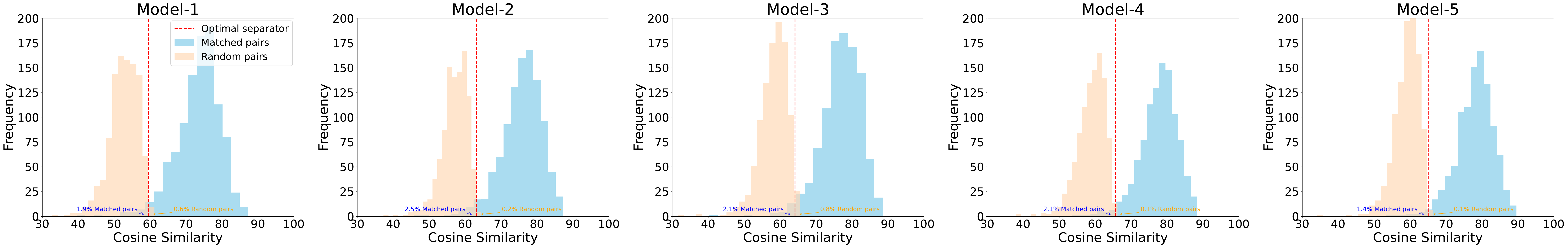}
        \caption*{(a) Polish-Ukrainian pair:  Ukrainian is $L_1$ ($s$), Polish is $L_2$ ($t$).}
    \end{subfigure}
  
  \vspace{0.5cm}

    \begin{subfigure}{\textwidth}
        \centering
        \includegraphics[width=\textwidth]{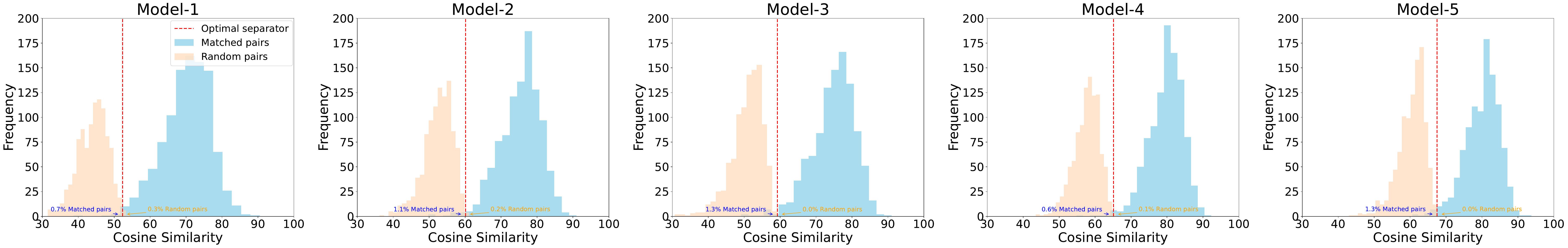}
        \caption*{(b) Hindi-Urdu pair: Urdu is $L_1$ ($s$), Hindi is $L_2$ ($t$).}
    \end{subfigure}

  \vspace{0.3cm}
  
  \caption{Histograms of similarities for matched sentence pairs and random pairs. Adding transliterated data in pretraining improves the overall similarities for both matched and random pairs. Leveraging auxiliary objectives improves the model's ability to differentiate between matched sentence pairs from random sentence pairs.
  }
  \label{fig:histogram_translation_sim_sr_f}
\end{figure*}

\begin{figure*}
    \setlength{\belowcaptionskip}{-0.5cm}

    \begin{subfigure}{\textwidth}
        \centering
        \includegraphics[width=\textwidth]{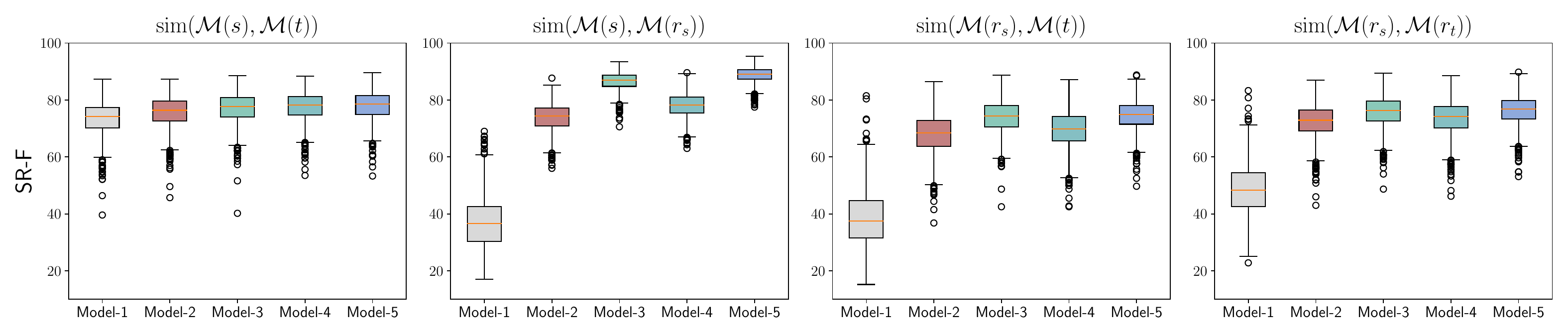}
        \caption*{(a) Polish-Ukrainian pair: Ukrainian is $L_1$ ($s$), Polish is $L_2$ ($t$).}
    \end{subfigure}
  
  \vspace{0.5cm}

    \begin{subfigure}{\textwidth}
        \centering
        \includegraphics[width=\textwidth]{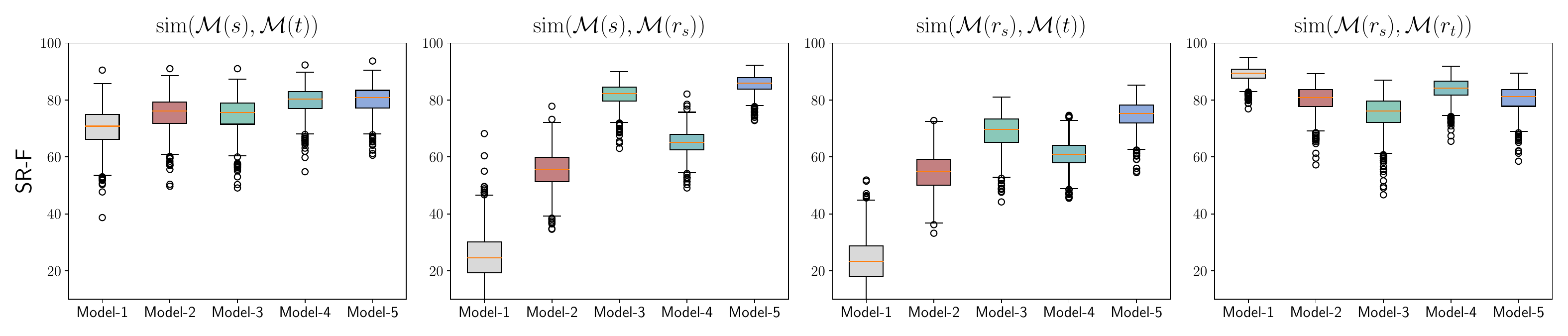}
        \caption*{(b) Hindi-Urdu pair: Urdu is $L_1$ ($s$), Hindi is $L_2$ ($t$).}
    \end{subfigure}
  \vspace{0.1cm}
  
  \caption{Comparison of different types of similarities (measured using \textbf{SR-F}).
  }
  \label{fig:sim_sr_f}
\end{figure*}

We show the similarity between matched sentence pairs and between random sentence pairs from SR-F in Figure~\ref{fig:histogram_translation_sim_sr_f}. We see a similar trend as for SR-B. However, because each sentence in Flores \cite{costa2022no} is relatively simple and quite different from the other sentences in the dataset, the similarity gap between matched and random pairs is already quite large. Therefore, the effect of including transliterations or auxiliary objectives is marginal.
~
Similarly, we visualize the four types of similarities in each model for SR-F in Figure \ref{fig:sim_sr_f}. The trend remains almost the same as for SR-B: including the transliterated data improves all similarities and the usage of transliteration-based alignment objectives can further improve overall similarities. 

\begin{figure*}
    \centering
        \setlength{\belowcaptionskip}{-0.4cm}

    \begin{subfigure}{\textwidth}
        \centering
    \includegraphics[width=0.188\textwidth]{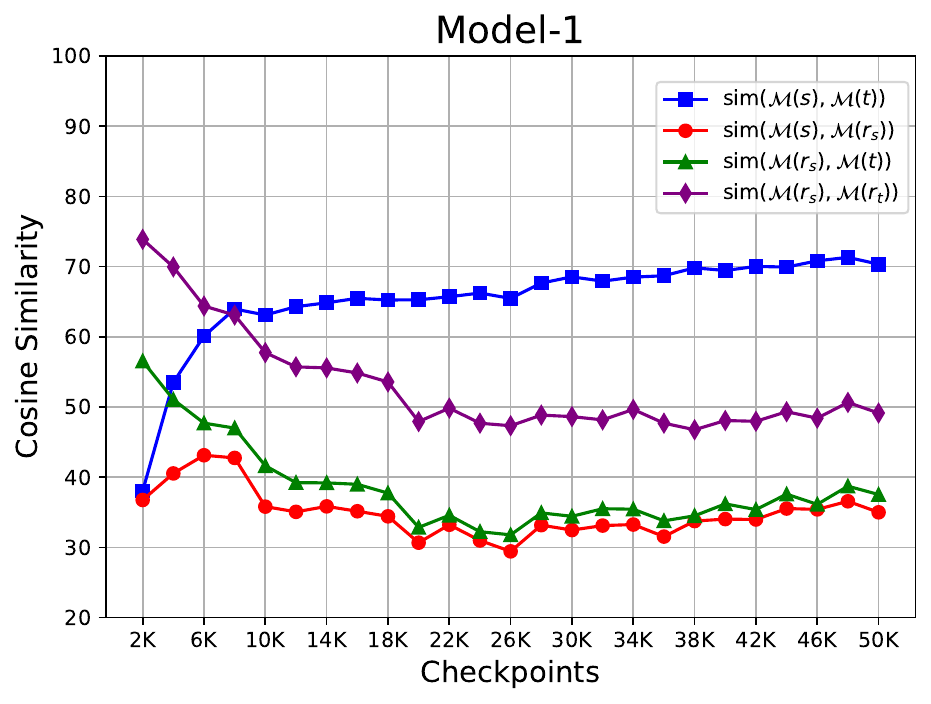}
    \includegraphics[width=0.19\textwidth]{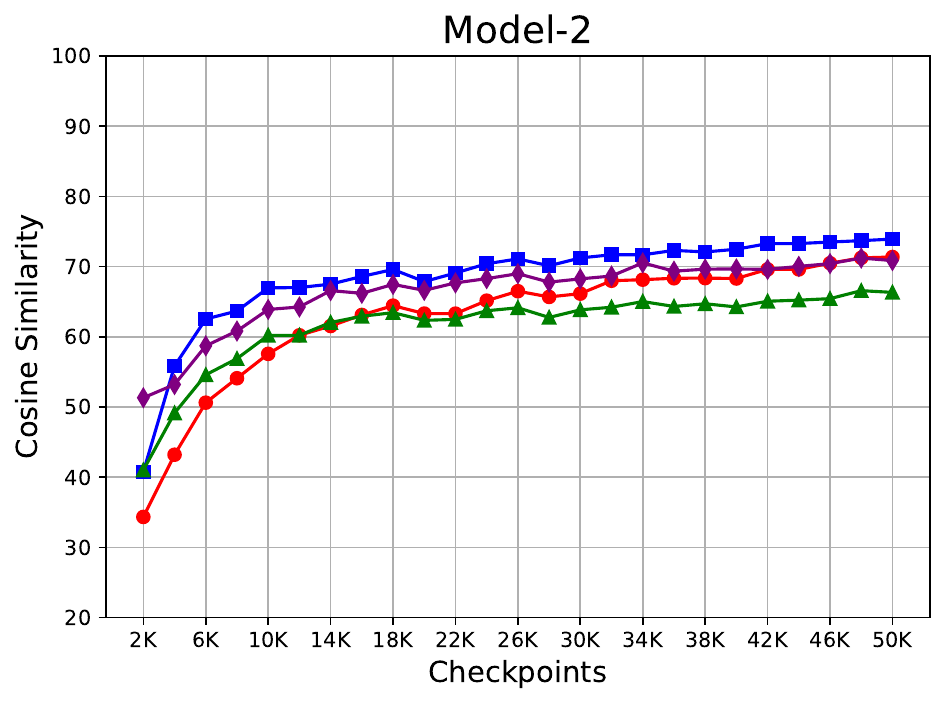}
    \includegraphics[width=0.19\textwidth]{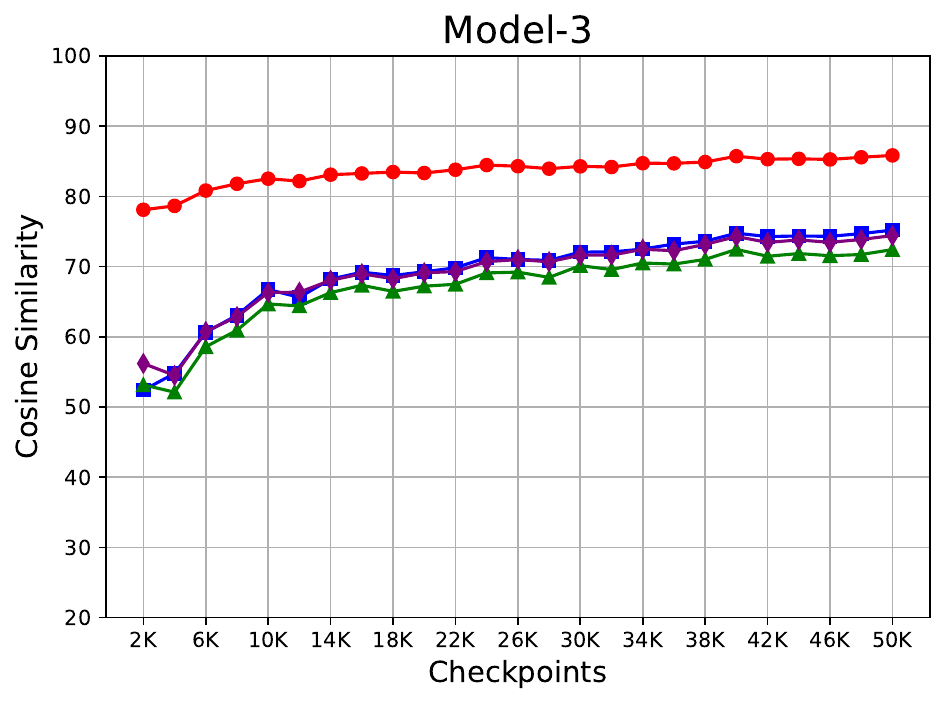}
    \includegraphics[width=0.19\textwidth]{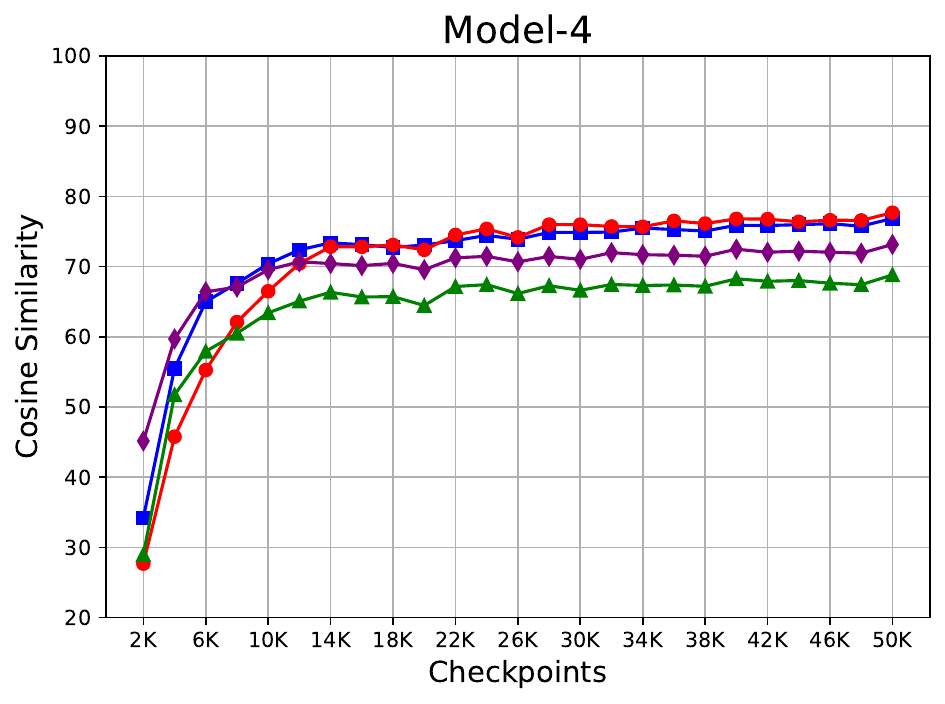}
    \includegraphics[width=0.19\textwidth]{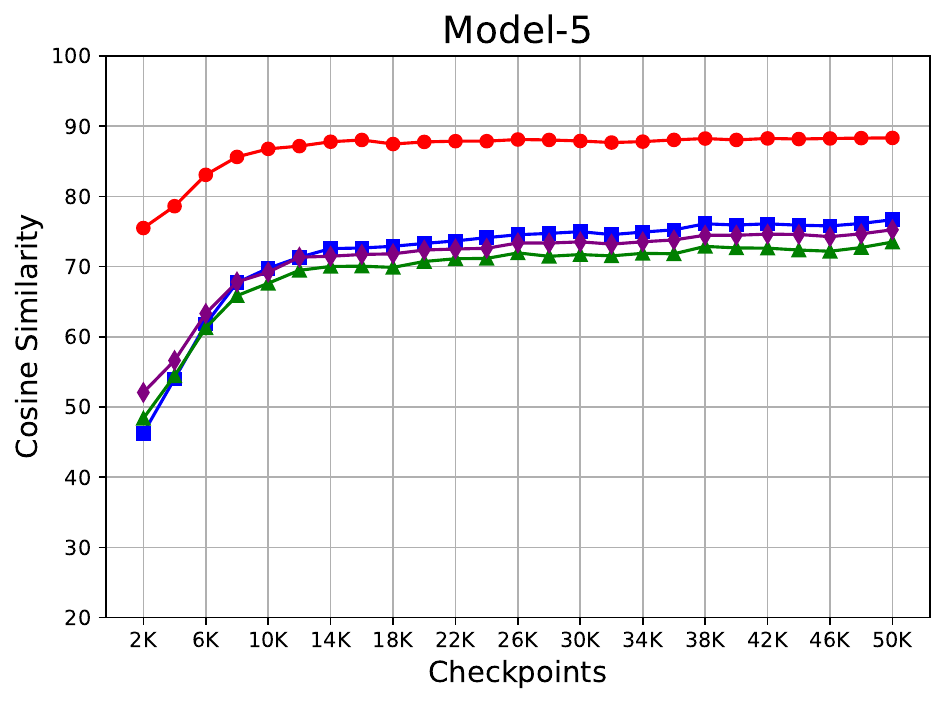}
    \caption*{(a) Polish-Ukrainian pair: Ukrainian is $L_1$ ($s$), Polish is $L_2$ ($t$).}
    \end{subfigure}
  
  \vspace{0.4cm}

    \begin{subfigure}{\textwidth}
        \centering
    \includegraphics[width=0.188\textwidth]{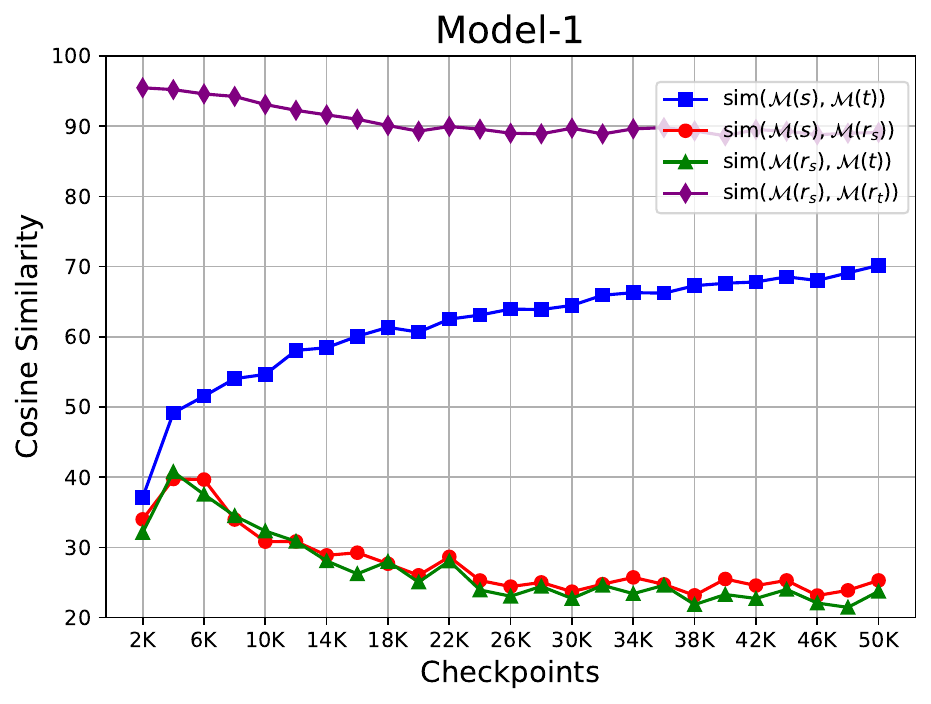}
    \includegraphics[width=0.19\textwidth]{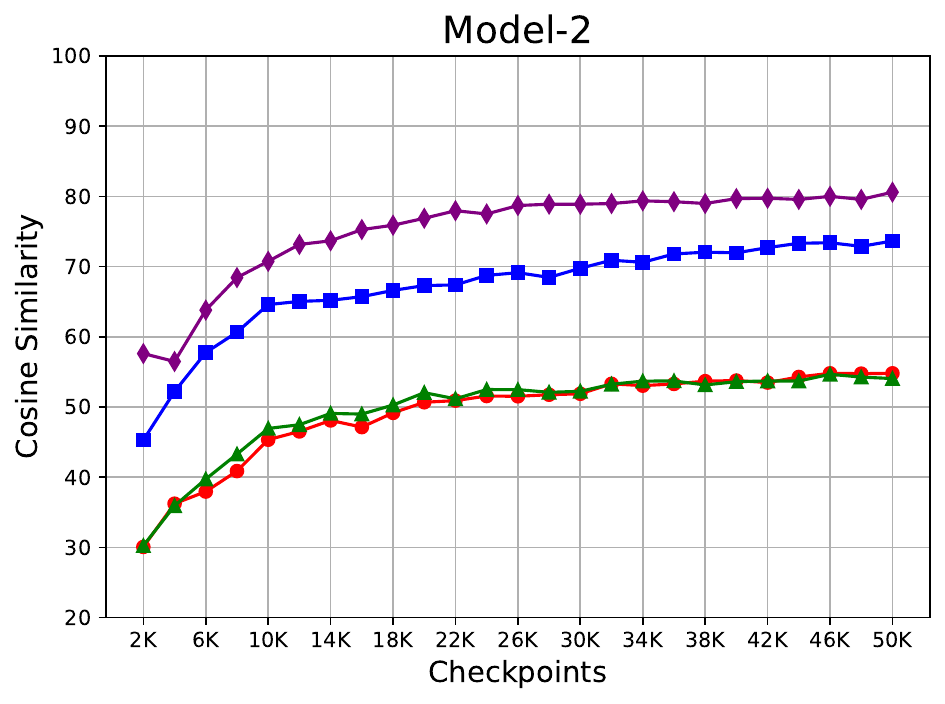}
    \includegraphics[width=0.19\textwidth]{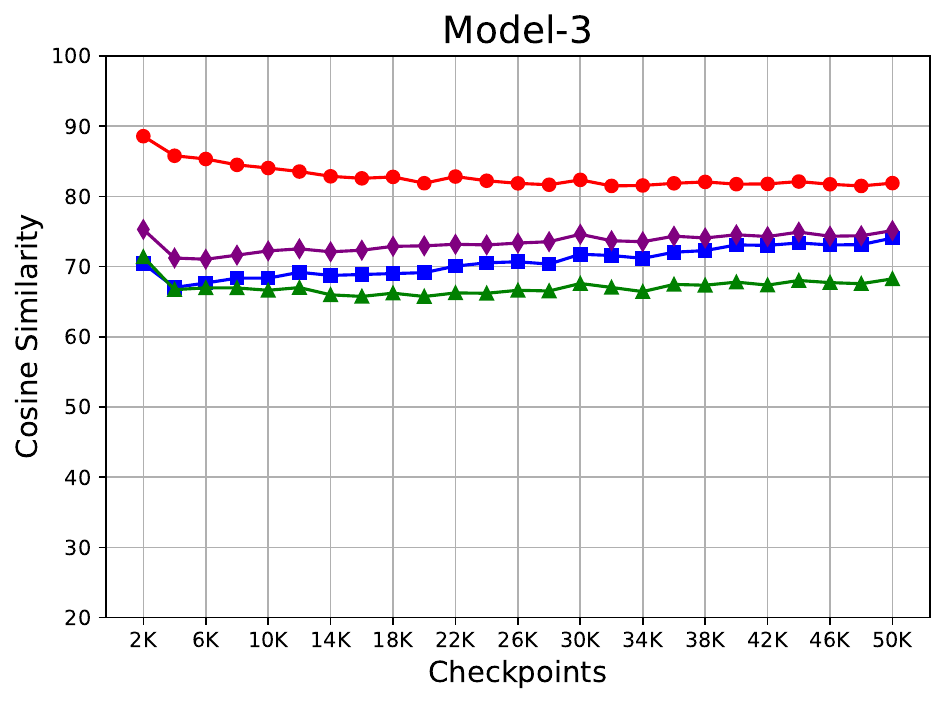}
    \includegraphics[width=0.19\textwidth]{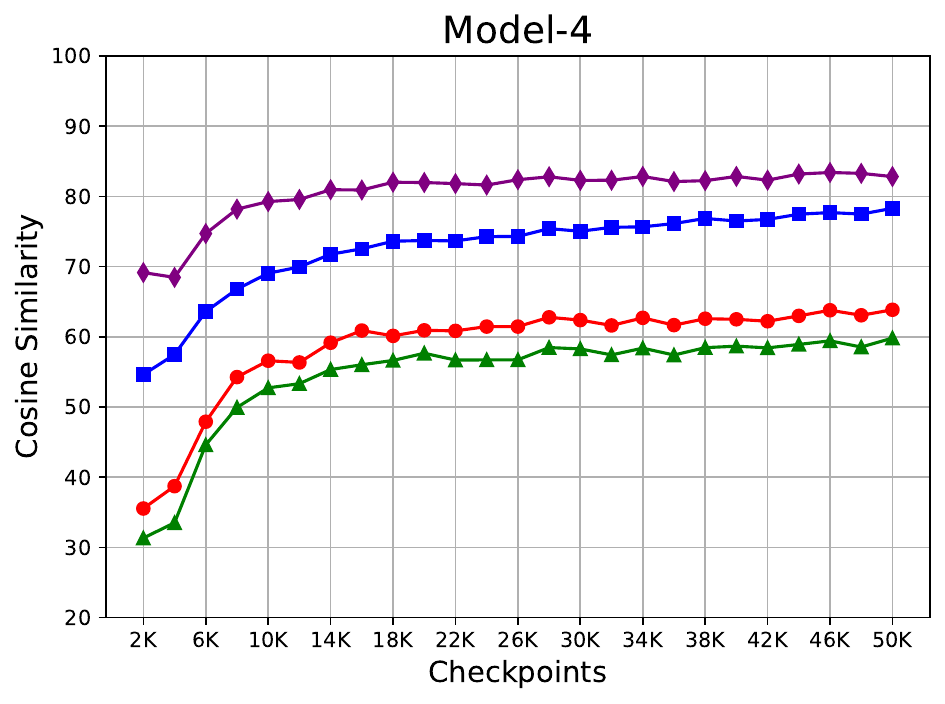}
    \includegraphics[width=0.19\textwidth]{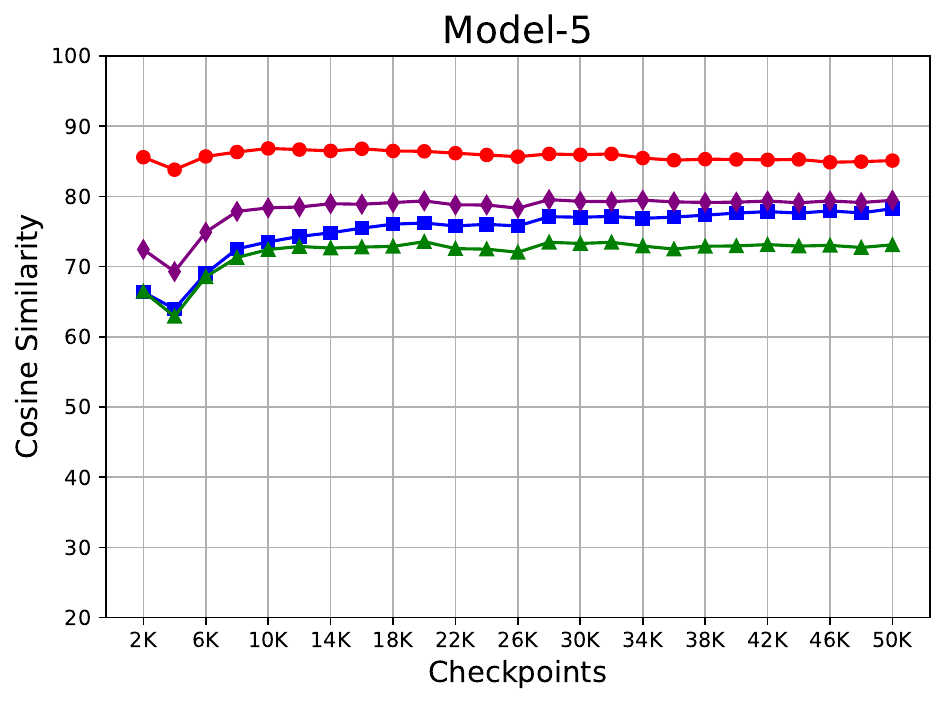}
        \caption*{(b) Hindi-Urdu pair: Urdu is $L_1$ ($s$), Hindi is $L_2$ ($t$).}
    \end{subfigure}

    \vspace{0.1cm}
    
    \caption{Dynamics of four types of similarities during training progression (measured using \textbf{SR-F}).}\label{fig:progression_sr_f}
\end{figure*}

We plot all four types of similarities measured using SR-F for each model and language pair throughtout pretraining in Figure \ref{fig:progression_sr_f}. The trend is almost identical to what we observe when measuring the similarity using SR-B: including transliterations has a direct effect on \textit{transliteration-transliteration} similarity and transliterations can implicitly improve the \textit{translation similarity} since transliterations work as an intermediary.

\section{Additional Analysis on the Other Direction}
\begin{figure*}
    \begin{subfigure}{\textwidth}
        \centering
        \includegraphics[width=\textwidth]{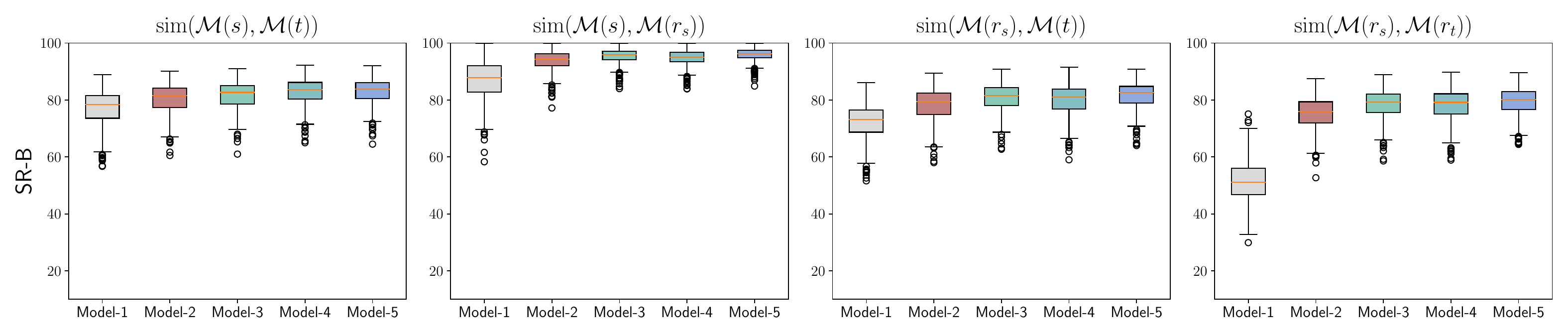}
        \caption*{(a) Polish-Ukrainian pair: Polish is $L_1$ ($s$), Ukrainian is $L_2$ ($t$)}
    \end{subfigure}
  
  \vspace{0.5cm}

    \begin{subfigure}{\textwidth}
        \centering
        \includegraphics[width=\textwidth]{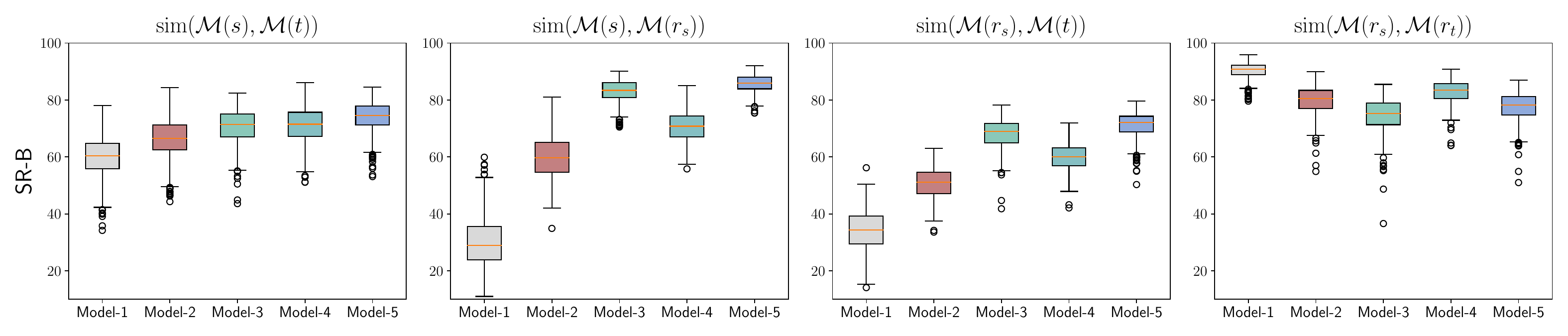}
        \caption*{(b) Hindi-Urdu pair: Hindi is $L_1$ ($s$), Urdu is $L_2$ ($t$).}
    \end{subfigure}
  \vspace{0.1cm}
  
  \caption{Comparison of different types of similarities for directions pol $\rightarrow$ ukr and hin $\rightarrow$ urd (measured using \textbf{SR-B}).
  }
  \label{fig:sim_sr_b_interchange}
\end{figure*}

\begin{figure*}
    \begin{subfigure}{\textwidth}
        \centering
        \includegraphics[width=\textwidth]{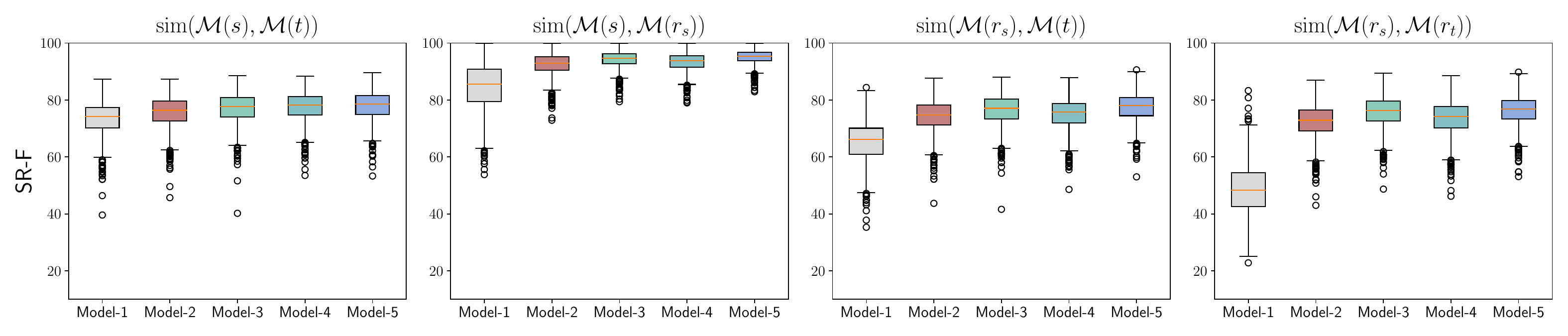}
        \caption*{(a) Polish-Ukrainian pair: Polish is $L_1$ ($s$), Ukrainian is $L_2$ ($t$).}
    \end{subfigure}
  
  \vspace{0.5cm}

    \begin{subfigure}{\textwidth}
        \centering
        \includegraphics[width=\textwidth]{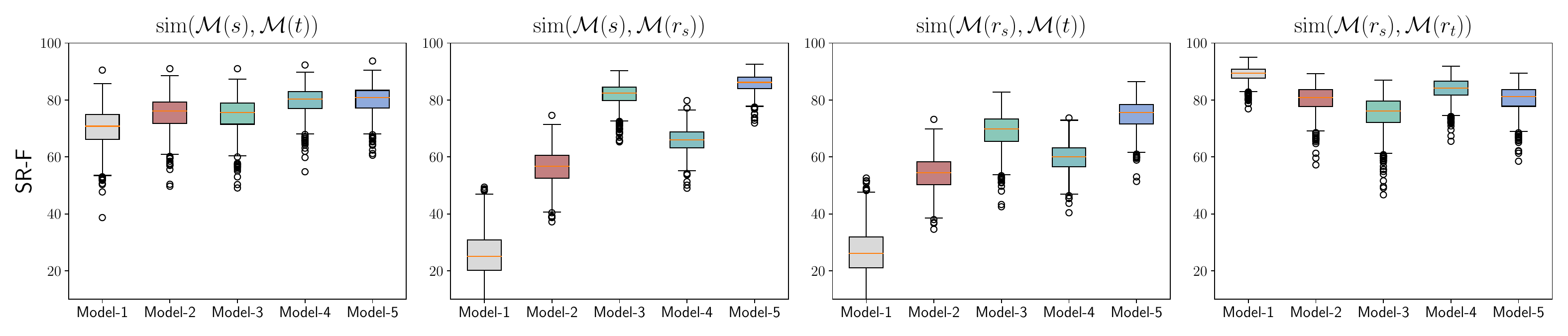}
        \caption*{(b) Hindi-Urdu pair: Hindi is $L_1$ ($s$), Urdu is $L_2$ ($t$).}
    \end{subfigure}
  \vspace{0.1cm}
  
  \caption{Comparison of different types of similarities for directions pol $\rightarrow$ ukr and hin $\rightarrow$ urd (measured using \textbf{SR-F}).
  }
  \label{fig:sim_sr_f_interchange}
\end{figure*}

\begin{figure*}
    \centering
        \setlength{\belowcaptionskip}{-0.4cm}

    \begin{subfigure}{\textwidth}
        \centering
    \includegraphics[width=0.188\textwidth]{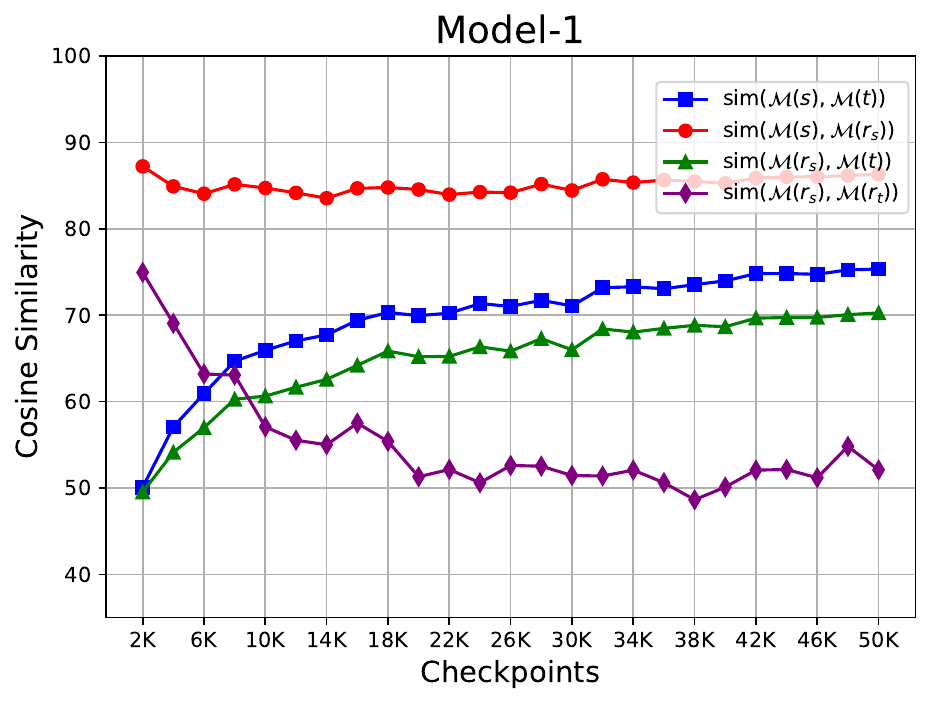}
    \includegraphics[width=0.19\textwidth]{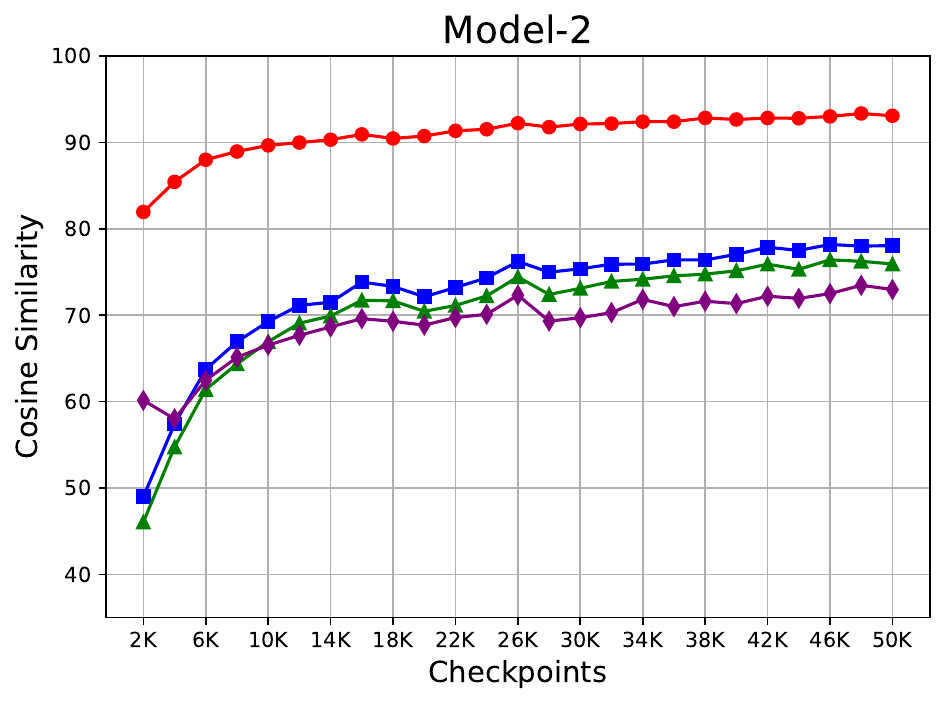}
    \includegraphics[width=0.19\textwidth]{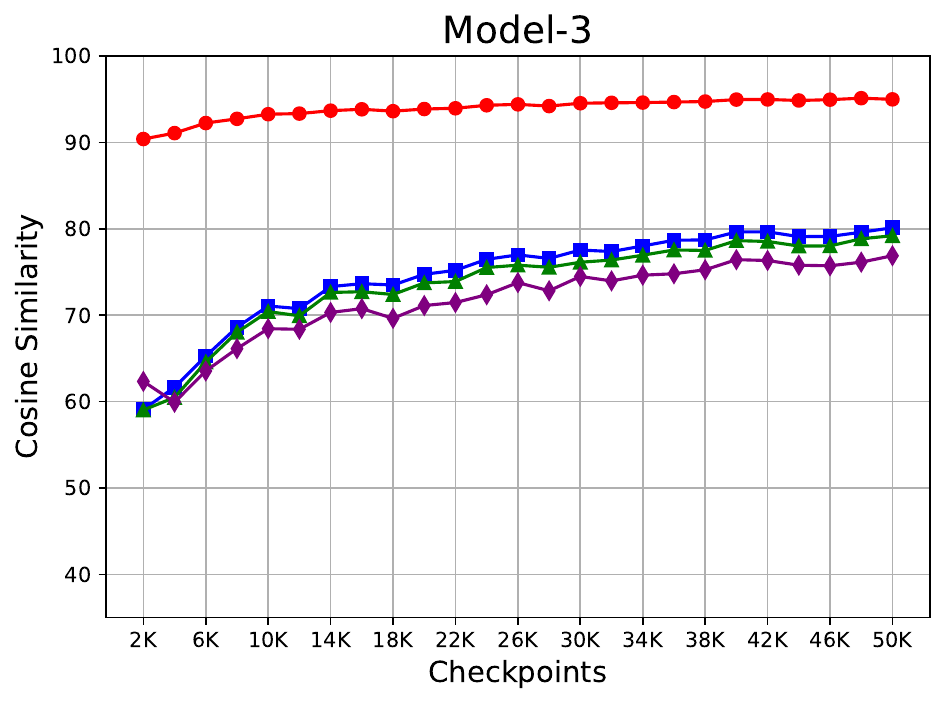}
    \includegraphics[width=0.19\textwidth]{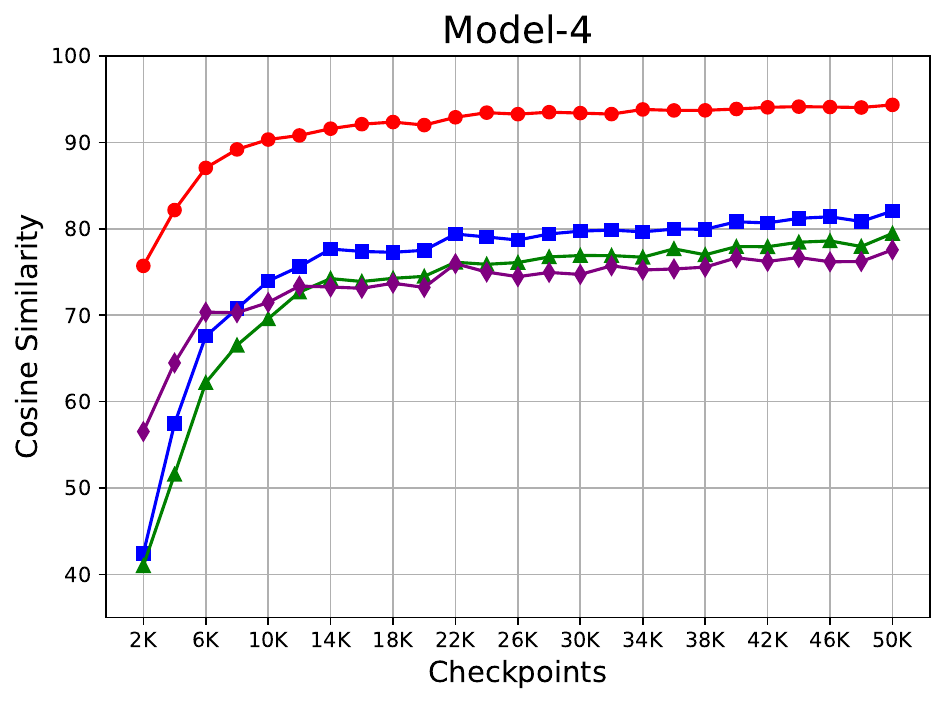}
    \includegraphics[width=0.19\textwidth]{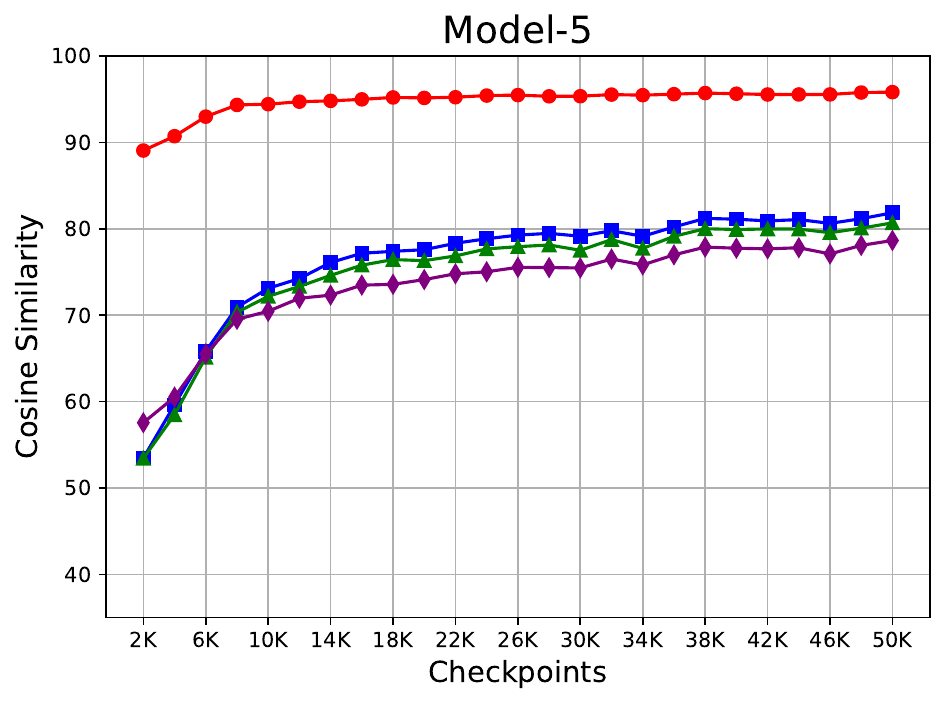}
    \caption*{(a) Polish-Ukrainian pair: Polish is $L_1$ ($s$), Ukrainian is $L_2$ ($t$).}
    \end{subfigure}
  
  \vspace{0.4cm}

    \begin{subfigure}{\textwidth}
        \centering
    \includegraphics[width=0.188\textwidth]{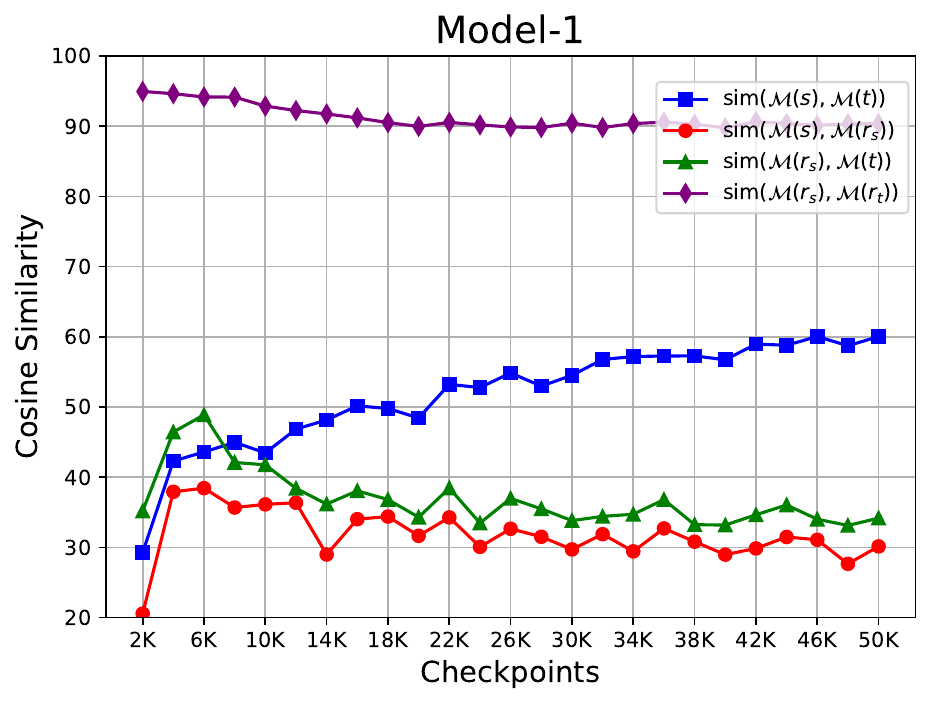}
    \includegraphics[width=0.19\textwidth]{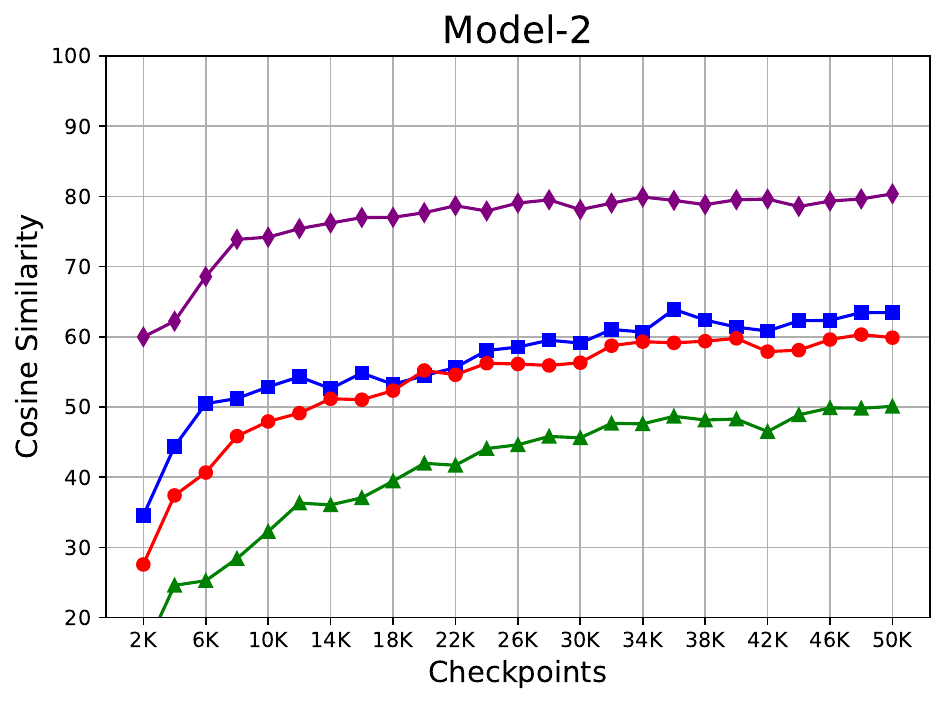}
    \includegraphics[width=0.19\textwidth]{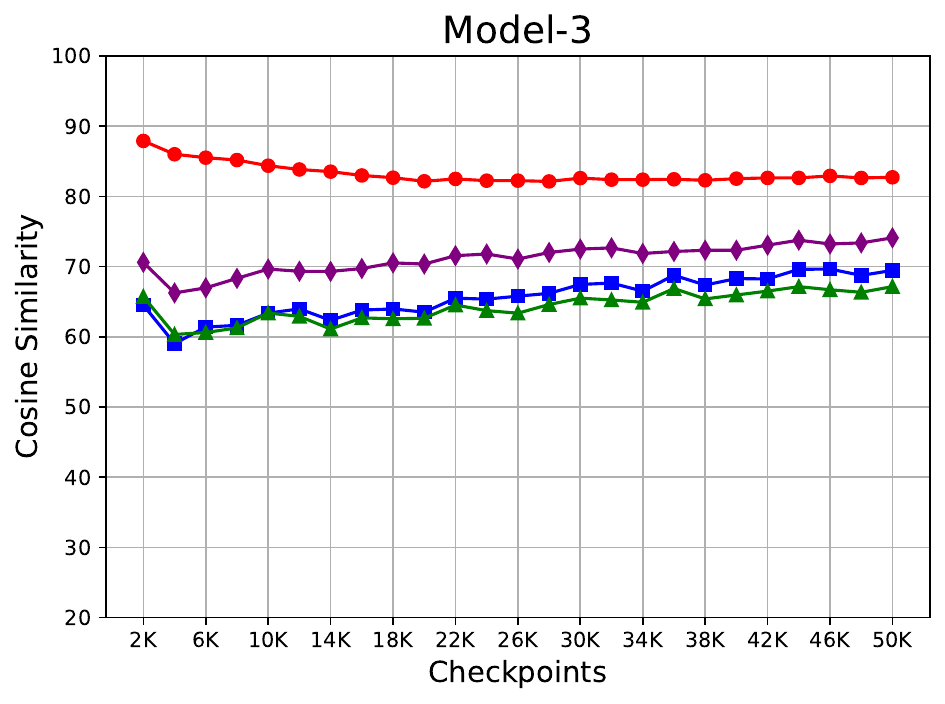}
    \includegraphics[width=0.19\textwidth]{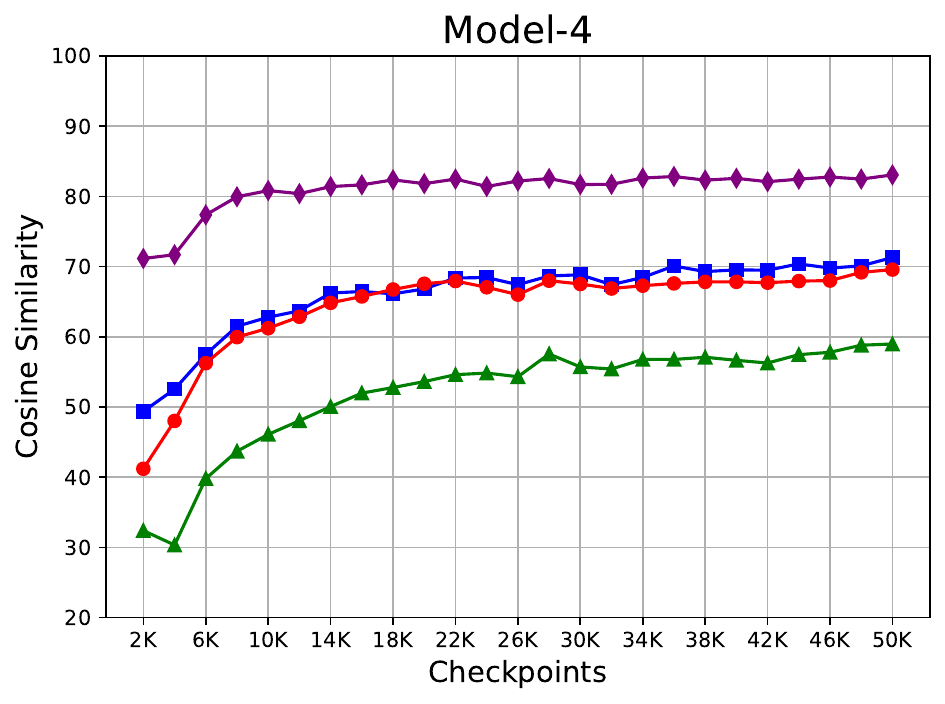}
    \includegraphics[width=0.19\textwidth]{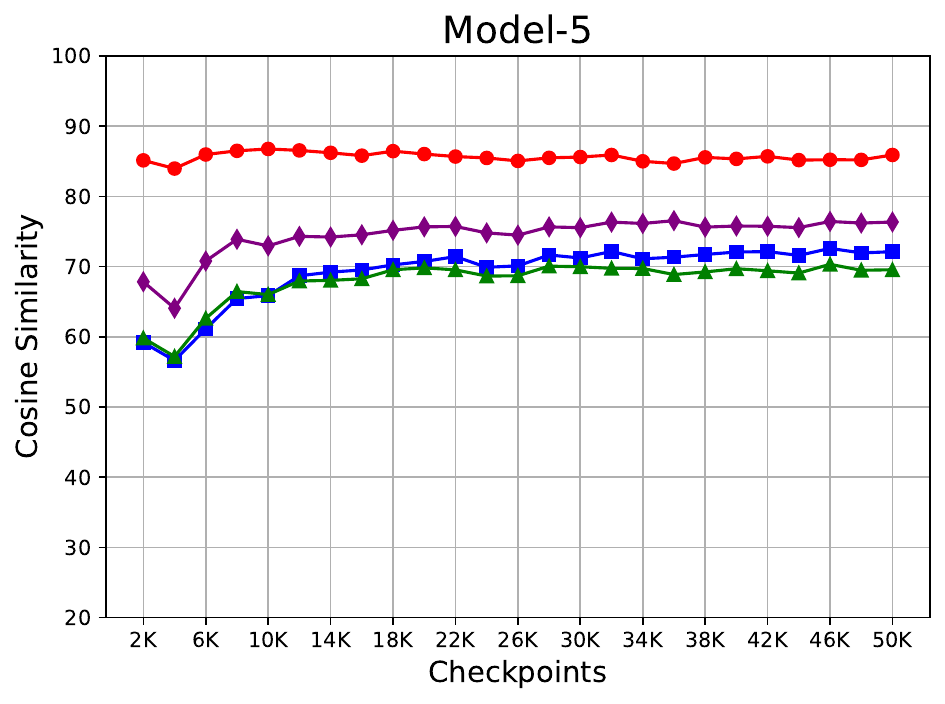}
        \caption*{(b) Hindi-Urdu pair: Hindi is $L_1$ ($s$), Urdu is $L_2$ ($t$).}
    \end{subfigure}

    \vspace{0.1cm}
    
    \caption{Dynamics of four types of similarities during training progression for directions pol $\rightarrow$ ukr and hin $\rightarrow$ urd (measured using \textbf{SR-B}).}
    \label{fig:progression_sr_b_interchange}
\end{figure*}

\begin{figure*}
    \centering
        \setlength{\belowcaptionskip}{-0.4cm}

    \begin{subfigure}{\textwidth}
        \centering
    \includegraphics[width=0.188\textwidth]{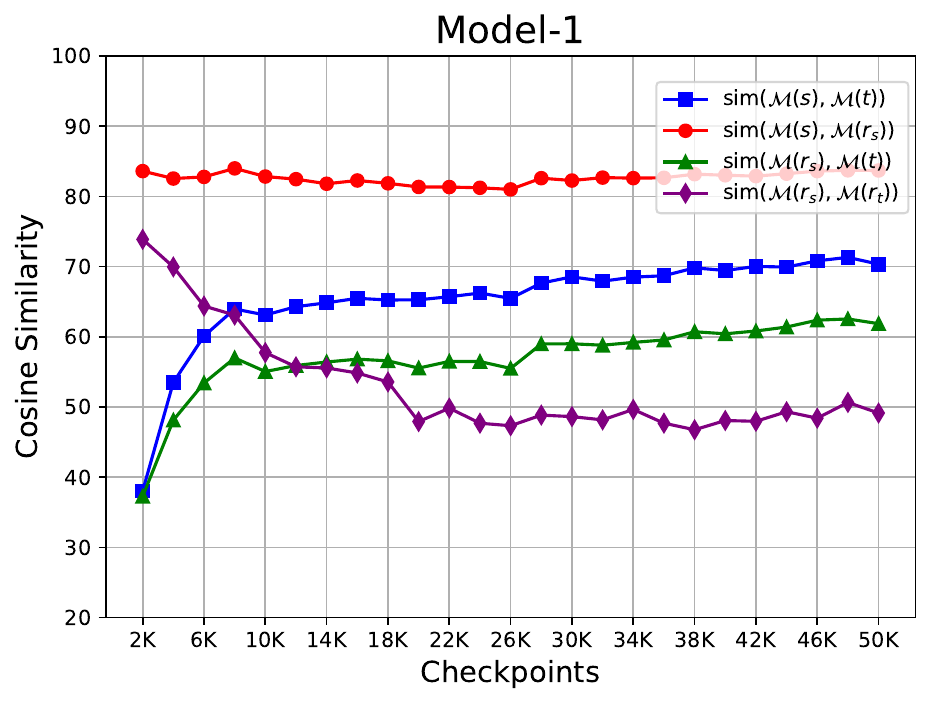}
    \includegraphics[width=0.19\textwidth]{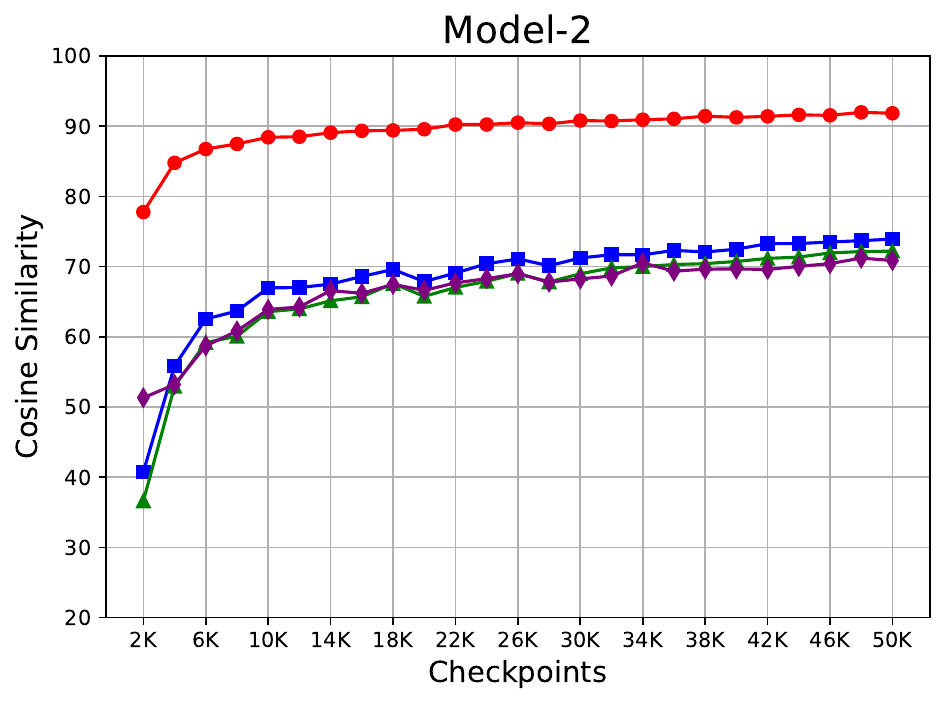}
    \includegraphics[width=0.19\textwidth]{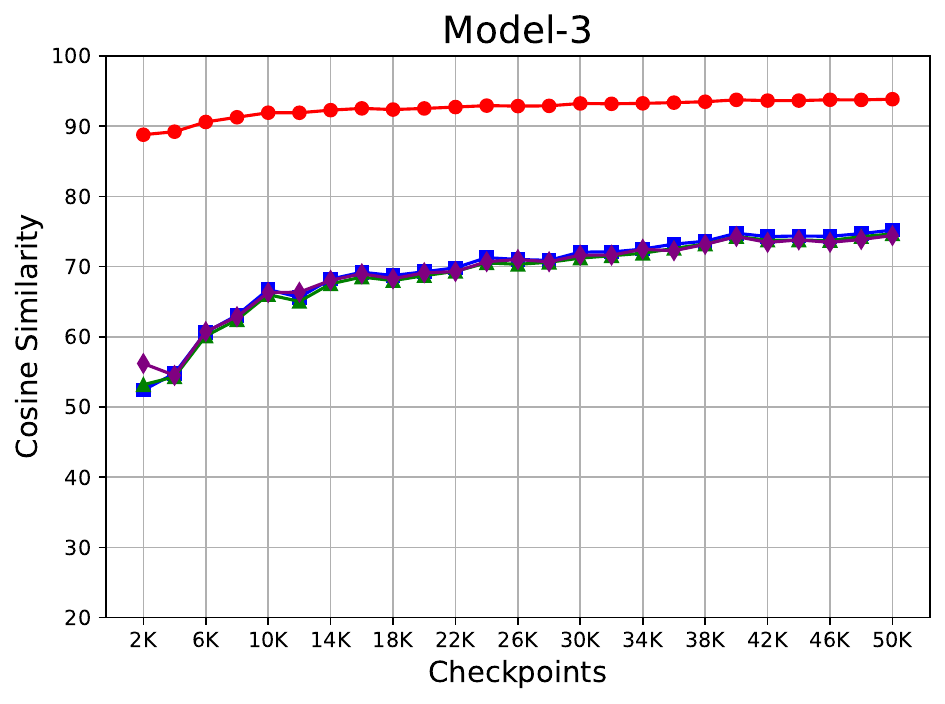}
    \includegraphics[width=0.19\textwidth]{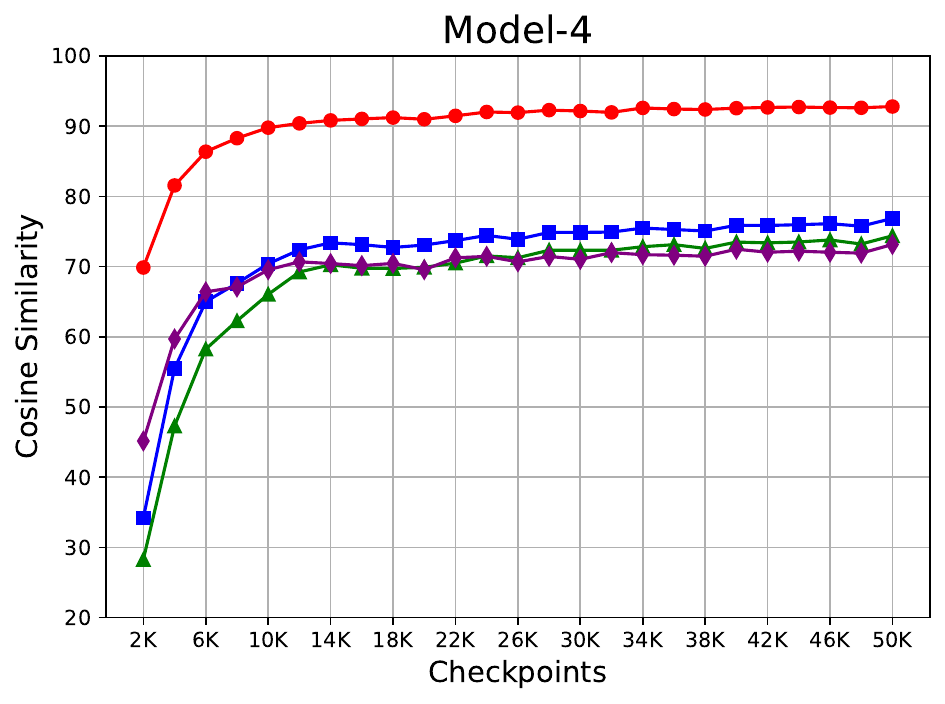}
    \includegraphics[width=0.19\textwidth]{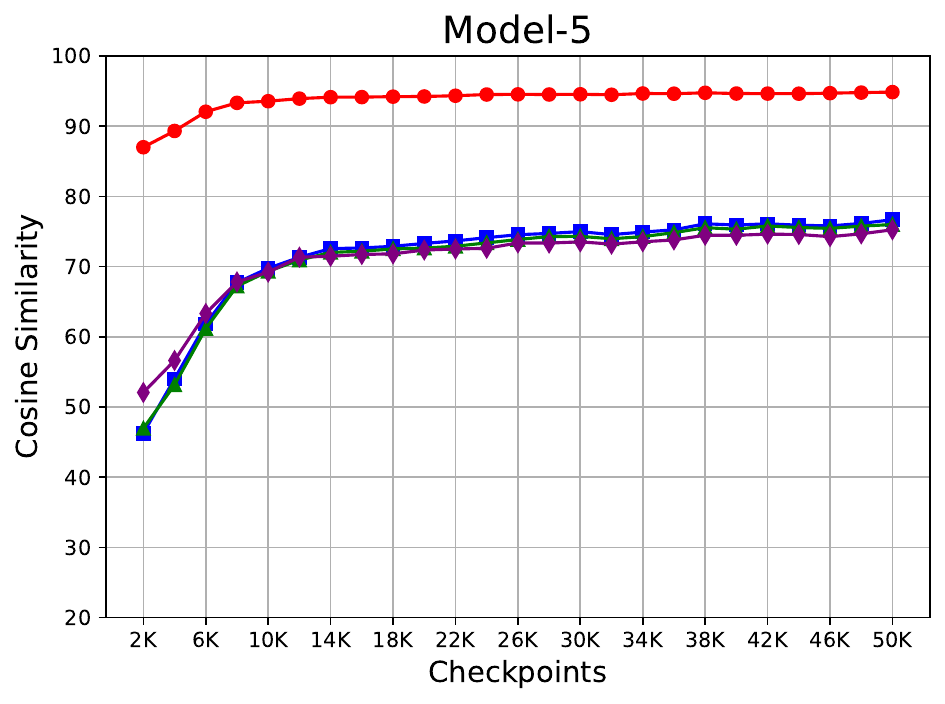}
    \caption*{(a) Polish-Ukrainian pair: Polish is $L_1$ ($s$), Ukrainian is $L_2$ ($t$).}
    \end{subfigure}
  
  \vspace{0.4cm}

    \begin{subfigure}{\textwidth}
        \centering
    \includegraphics[width=0.188\textwidth]{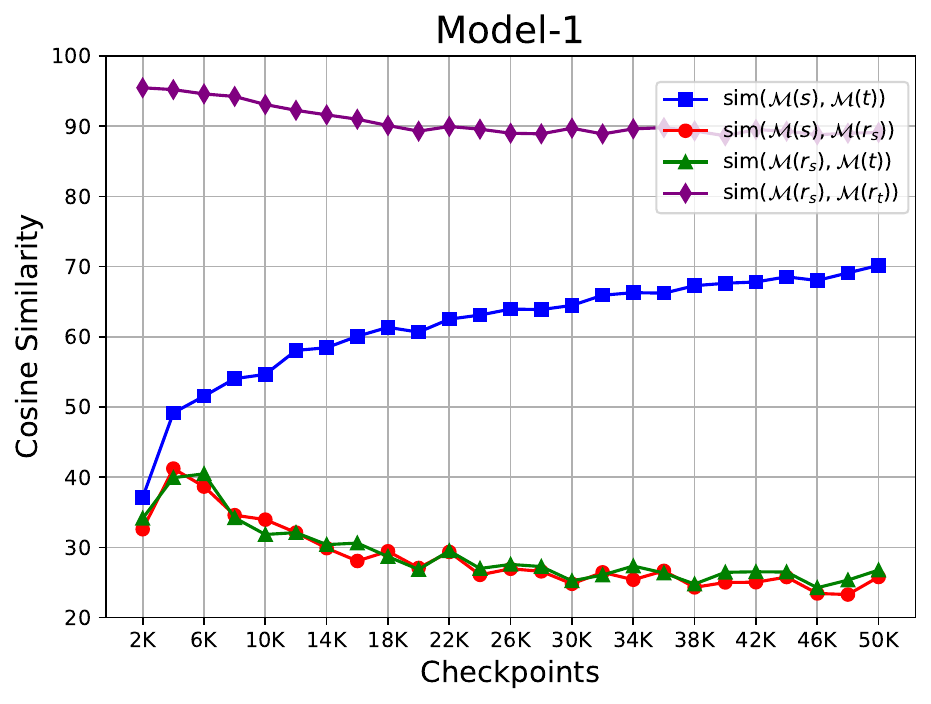}
    \includegraphics[width=0.19\textwidth]{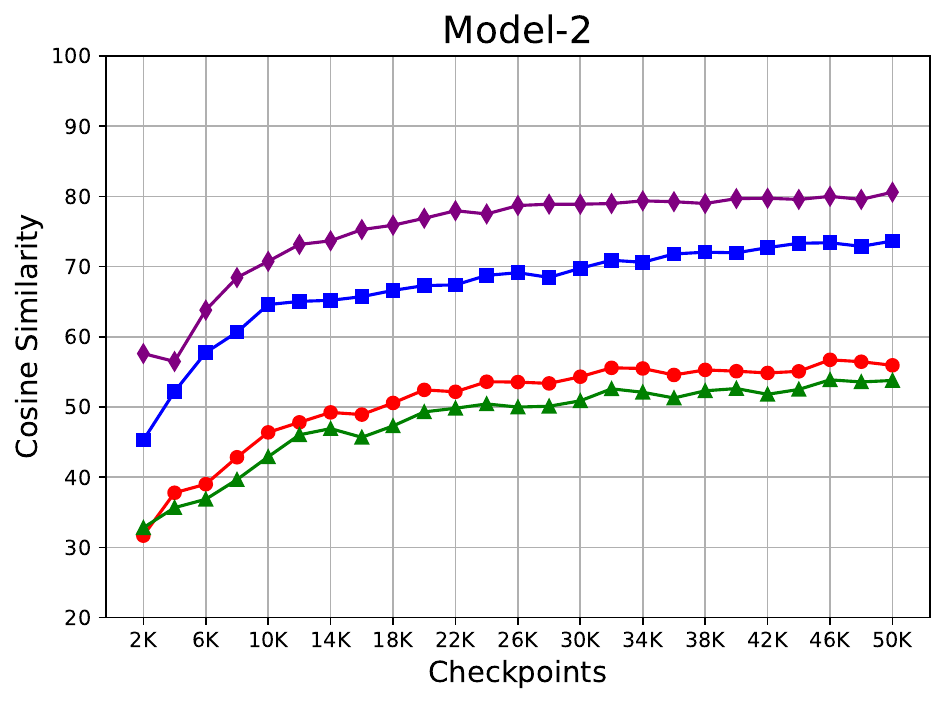}
    \includegraphics[width=0.19\textwidth]{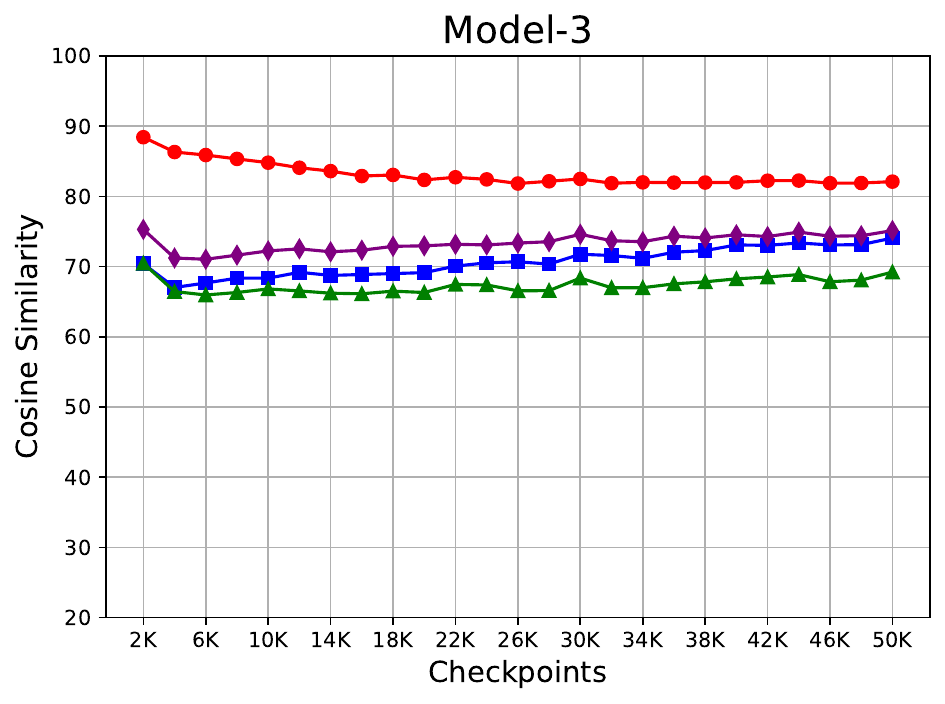}
    \includegraphics[width=0.19\textwidth]{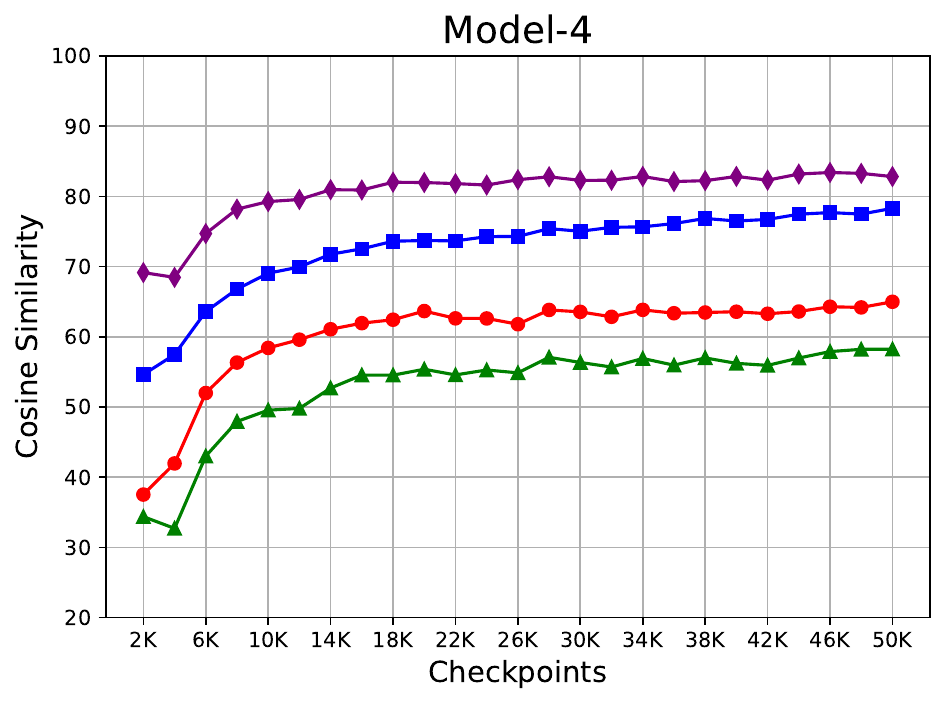}
    \includegraphics[width=0.19\textwidth]{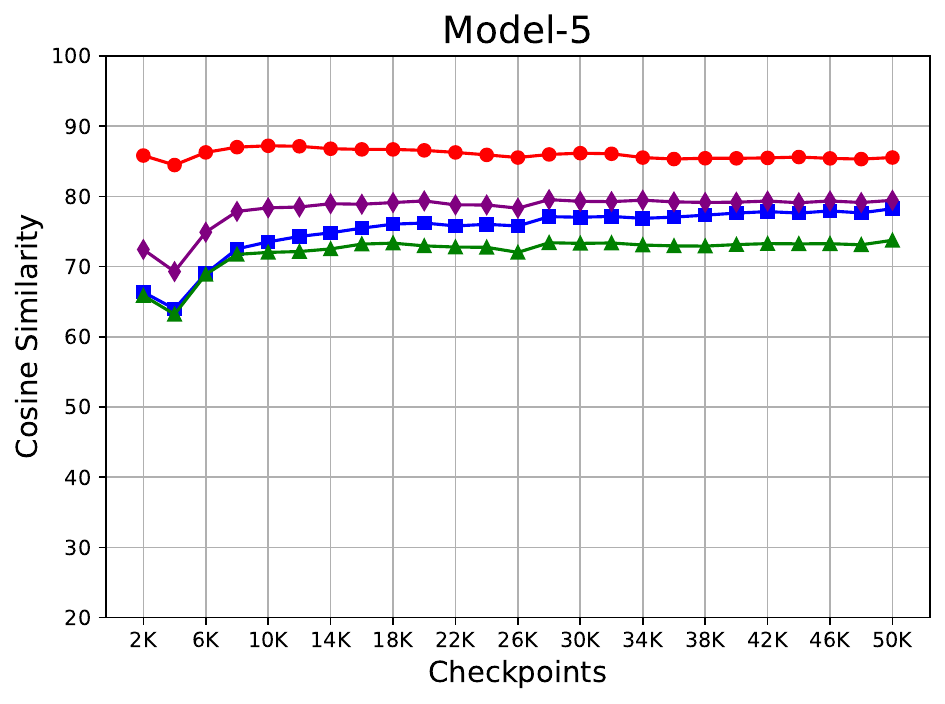}
        \caption*{(b) Hindi-Urdu pair: Hindi is $L_1$ ($s$), Urdu is $L_2$ ($t$).}
    \end{subfigure}

    \vspace{0.1cm}
    
    \caption{Dynamics of four types of similarities during training progression for directions pol $\rightarrow$ ukr and hin $\rightarrow$ urd (measured using \textbf{SR-F}).}\label{fig:progression_sr_f_interchange}
\end{figure*}

\seclabel{additonal_direction}
We also compute the different types of similarities using the other directions for each language pair. Specifically, we use pol $\rightarrow$ ukr for the pol-ukr pair and hin $\rightarrow$ urd for the hin-urd pair. We show the comparison of the similarities in Figure \ref{fig:sim_sr_b_interchange} (using SR-B) and in Figure \ref{fig:sim_sr_f_interchange} (using SR-F). Additionally, we show the dynamics of how the similarities vary throughout the pretraining phase in Figure \ref{fig:progression_sr_b_interchange} (using SR-B) and Figure \ref{fig:progression_sr_f_interchange} (using SR-F).

The general trend remains roughly the same for the hin-urd pair regardless of which direction is used for calculating the similarity. For the pol-ukr pair, because Polish uses Latin script by default and Uroman only removes the diacritics, the \textit{transliteration similarity}, i.e.,  $\text{sim}(\mathcal{M}(s), \mathcal{M}(r_s))$, remains high throughout the pretraining, as shown in Figure \ref{fig:progression_sr_b_interchange} and Figure \ref{fig:progression_sr_f_interchange}. We also observe that, without including transliterated data (Model-1), the model already yields high \textit{transliteration similarity}. Once the transliterated data is included in the pretraining (Model-2, -3, -4, and -5), the \textit{transliteration similarity} further improves, as shown in Figure \ref{fig:sim_sr_b_interchange} and Figure \ref{fig:sim_sr_f_interchange}, which is expected.

\end{document}